\documentclass[letterpaper, 10 pt, conference]{ieeeconf}  
\IEEEoverridecommandlockouts
\usepackage{multirow}
\usepackage{graphicx}
\usepackage{subcaption}
\usepackage{amsmath} 
\usepackage{amssymb}
\let\labelindent\relax
\usepackage{enumitem}
\usepackage{url}

\title{\LARGE \bf
Bimanual Robot-Assisted Dressing: A Spherical Coordinate-Based Strategy for Tight-Fitting Garments}

\author{Jian Zhao$^{1}$, Yunlong Lian$^{1}$, Andy M Tyrrell$^{1}$, Michael Gienger$^{2}$, and Jihong Zhu$^{1}$
\thanks{$^{1}$Jian Zhao, Yunlong Lian, Andy Tyrrell and Jihong Zhu are with School of Physics, Engineering and Technology, University of York, UK. Corresponding author:
        {\tt\small jihong.zhu@york.ac.uk}}%
\thanks{$^{2}$Michael Gienger is  with Honda Research Institute  Europe, Germany.
        }%
}

\begin{document}

\maketitle
\thispagestyle{empty}
\pagestyle{empty}

\begin{abstract}
Robot-assisted dressing is a popular but challenging topic in the field of robotic manipulation, offering significant potential to improve the quality of life for individuals with mobility limitations. Currently, the majority of research on robot-assisted dressing focuses on how to put on loose-fitting clothing, with little attention paid to tight garments. For the former, since the armscye is larger, a single robotic arm can usually complete the dressing task successfully. However, for the latter, dressing with a single robotic arm often fails due to the narrower armscye and the property of diminishing rigidity in the armscye, which eventually causes the armscye to get stuck. This paper proposes a bimanual dressing strategy suitable for dressing tight-fitting clothing. To facilitate the encoding of dressing trajectories that adapt to different human arm postures, a spherical coordinate system for dressing is established. We uses the azimuthal angle of the spherical coordinate system as a task-relevant feature for bimanual manipulation. Based on this new coordinate, we employ Gaussian Mixture Model (GMM) and Gaussian Mixture Regression (GMR) for imitation learning of bimanual dressing trajectories, generating dressing strategies that adapt to different human arm postures. The effectiveness of the proposed method is validated through various experiments.
\end{abstract}

\section{INTRODUCTION}
\label{sec:intro}

With the increasing severity of population aging, how to provide high-quality and long-term care for the elderly has become a challenging global issue. According to data from the World Health Organization, the proportion of the world's population aged 60 and over will nearly double from 12\% to 22\% between 2015 and 2050, and by 2050, 80\% of the elderly will be living in low- and middle-income countries \cite{who2021ageing}. Assistive robots can help the elderly and disabled with daily life tasks and are potentially more cost-effective than human labor, making them a promising alternative to the shortage of caregiving personnel. Among various daily care tasks, assisting with dressing has been reported as one of the most labor-intensive and least automated tasks \cite{dudgeon2008managing}. 

\begin{figure}[tbp]
  \centering
  \begin{subfigure}[b]{0.23\textwidth}
    \includegraphics[width=\textwidth]{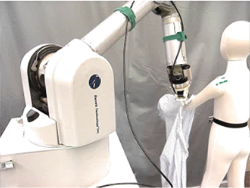}
    \caption{}
    \label{fig:demo_dressing_tasks:demo_a}
  \end{subfigure}
  \hfill 
  \begin{subfigure}[b]{0.23\textwidth}
    \includegraphics[width=\textwidth]{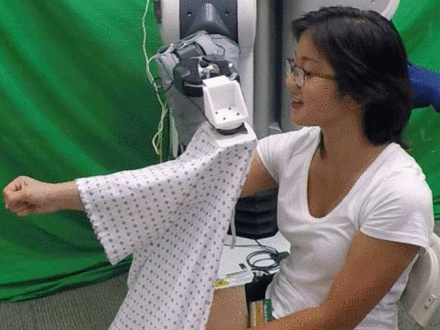}
    \caption{}
    \label{fig:demo_dressing_tasks:demo_b}
  \end{subfigure}
  \vspace{0.5cm} 
  \begin{subfigure}[b]{0.23\textwidth}
    \includegraphics[width=\textwidth]{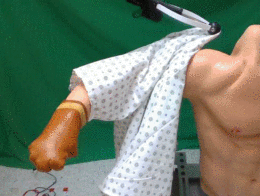}
    \caption{}
    \label{fig:demo_dressing_tasks:demo_c}
  \end{subfigure}
  \hfill 
  \begin{subfigure}[b]{0.23\textwidth}
    \includegraphics[width=\textwidth]{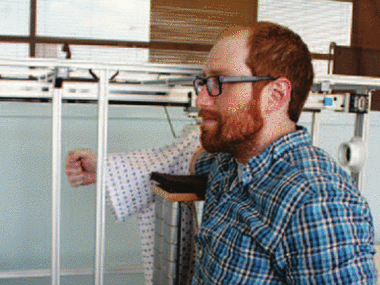}
    \caption{}
    \label{fig:demo_dressing_tasks:demo_d}
  \end{subfigure}
  \caption{Example robot-assisted dressing tasks, all of which use loose hospital gowns. (a) demo A \cite{hoyos2016incremental}. (b) demo B \cite{Zackory2018Tracking}. (c) demo C \cite{Yufei2022Visual}. (d) demo D \cite{Wenhao2017Haptic}.}
  \vspace{-8mm}
  \label{fig:demo_dressing_tasks}
\end{figure}

There has been research into robot-assisted dressing (see Fig. \ref{fig:demo_dressing_tasks}). However, most of these studies focus on strategies for dressing loose-fitting clothing, with little attention has been paid to tight-fitting garments. In this paper, the terms \textquotedblleft loose\textquotedblright\ and \textquotedblleft tight\textquotedblright\ refer to the ratio of the circumference of the armscye (i.e., armhole) to the human arm circumference. Examples of tight-fitting garments include compression shirts, fitted sweaters, and long-sleeve sportswear made from elastic materials. A larger ratio indicates a looser fit, while a smaller ratio indicates a tighter fit. For loose clothing, its characteristic is a larger armscye area, resulting in less contact with the human arm during the dressing process. Therefore, a single robotic arm is often sufficient to successfully complete the dressing task. However, when dressing with a single robotic arm in tight clothing, it often fails. This is because of the property of diminishing rigidity \cite{Dmitry2013Manipulation} in the armscye, which eventually causes the armscye to get stuck \cite{Jihong2024Do}. To address this problem, this paper proposes using bimanual robots to simultaneously dress tight-fitting clothing, ensuring that the clothing smoothly passes the human arm. Based on this idea, a learning-based bimanual dressing strategy is proposed.

\begin{figure}[tbp]
  \centering %
  \includegraphics[width=0.4\textwidth]{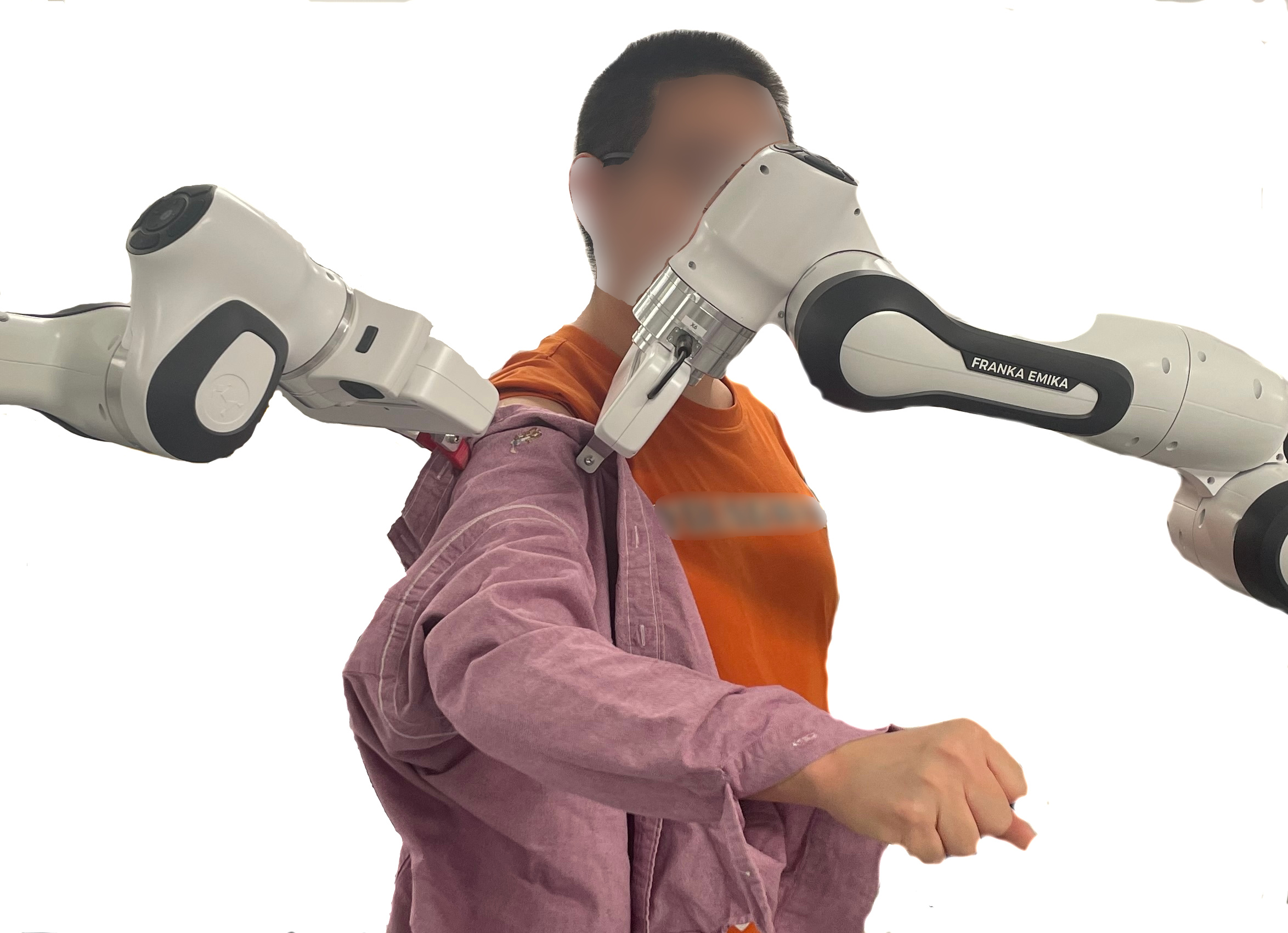} %
  \caption{Bimanual robot-assisted dressing setup. Two Franka Research 3 robotic arms on both sides of the human arm are deployed to assist together in dressing tight clothing.} %
  \label{fig:dressing_framework} %
   \vspace{-5mm}
\end{figure}

The proposed system setup is shown in Fig. \ref{fig:dressing_framework}. Two Franka Research 3 robotic arms are positioned on two sides of the human arm, working together to help the human put on the clothing. The dressing task involves the robot pulling a piece of clothing over a person's arm, with the dressing path starting from the wrist joint, passing through the elbow joint, and finally reaching the shoulder joint. We adopt a common assumption employed in previous related studies \cite{hoyos2016incremental, pignat2017learning, Jihong2022Learning}, where the human arm remains stationary during the dressing process. This approach avoids challenges such as visual occlusion \cite{Greg2018Elbows} and dynamic human arm posture estimation \cite{Fan2019Probabilistic}, allowing a focus on the study of the bimanual dressing strategies.

Since humans are inherently skilled at assisting in dressing, we use imitation learning to teach the robot dressing skills. To achieve the learning-based bimanual dressing task, the core problem to solve is identifying task-relevant features for bimanual manipulation \cite{edsinger2007two}. To address this issue, a spherical dressing coordinate system is estebished to better encode the dressing trajectory. After mapping the bimanual dressing trajectories to the spherical coordinate system, correlation between the azimuthal angles of the two spherical trajectories is observed. Azimuthal angle is used as the task-relevant feature for bimanual dressing. For each individual robotic arm, according to \cite{Jihong2024Do}, the key factor influencing the dressing strategy is the elbow joint angle. Considering this factor, GMM to learn from human demonstration trajectories and GMR to generate new trajectories are used. GMM is a probabilistic model that represents data as a mixture of multiple Gaussian distributions, while GMR is a regression technique derived from GMM that allows generating smooth output trajectories based on learned distributions.

The main contribution of this paper is proposing a bimanual dressing assistance method suitable for dressing tight-fitting clothing. To this end, a spherical dressing coordinate system is established to facilitate the encoding of bimanual dressing trajectories and uses the azimuthal angle as the task-relevant feature for bimanual dressing. A bimanual dressing strategy is then designed using a GMM/GMR-based imitation learning. While the task scenario emphasizes tight-fitting garments to highlight the challenges of close-contact manipulation, the primary focus of this work lies in the trajectory generation algorithm itself, which is generally applicable to other dressing tasks as well.

The remainder of this paper is organized as follows: Section \ref{sec:related_work} reviews the current state of research in the field of robot-assisted dressing. The detailed methodology of the approach is presented in Section \ref{sec:method}, including the establishment of the spherical coordinate system and the application of GMM/GMR in bimanual assisted dressing. Section \ref{sec:experiment} presents the experimental results and discusses the effectiveness of the proposed method. Finally, Section \ref{sec:conclusion} concludes the paper.

\section{RELATED WORK}
\label{sec:related_work}

\subsection{Robotic Dressing Assistance}
\label{subsec:related_work:robotic_dress_assist}

Robot-assisted dressing refers to robots helping humans put on various types of clothing, such as T-shirts\cite{Tomoya2011Reinforcement, joshi2019framework}, jackets\cite{pignat2017learning}, pants\cite{yamazaki2016bottom}, and shoes\cite{Gerard2018Joining,Aleksandar2019Personalized}, which is currently a popular topic. This paper primarily analyze research related to human arm dressing.

Imitation learning is a popular method for acquiring dressing strategies. Pignat and Calinon \cite{pignat2017learning} used a hidden semi-Markov model (HSMM) to encode the user's sensory information and the robot's motion commands as a joint distribution for jackets, enabling robot-assisted dressing through human demonstrations. Hoyos et al. \cite{hoyos2016incremental} applied incremental learning techniques to a task-parameterized Gaussian mixture (TP-GMM) model for loose-fitting medical long-sleeved shirts, allowing the addition of new trajectory information without retraining the entire task. Joshi et al. \cite{joshi2019framework} proposed a mixed method combining the dynamic movement primitives (DMP) and Bayesian Gaussian process latent variable model (BGPLVM) for sleeveless shirts to achieve tasks such as human arm dressing and human body dressing.

In addition to imitation learning, other methods have been applied to the field of robot-assisted dressing. Clegg et al. \cite{Alexander2020Learning} used deep reinforcement learning to accomplish a series of simulated tasks for dressing the sleeves of a hospital gown. Yu et al. \cite{Wenhao2017Haptic} used haptic sensor learning and employed an HMM to predict whether the task of dressing in loose robes was successful. Erickson et al. \cite{Zackory2018Tracking} used capacitive proximity sensors to track human pose and accomplish the task of dressing in loose robes.

Despite the significant amount of research into robot-assisted dressing, most studies focus on relatively loose-fitting clothing, with little attention paid to the challenges of dressing tight-fitting clothing. This is the starting point of the research presented in this paper.

\subsection{Bimanual Manipulation}
\label{subsec:related_work:bimanual_mani}

The research reported in this paper falls under the category of bimanual operations, where two robotic arms work together to assist a single human arm in dressing. We consider specifically the coordination between two arms which is classified as symmetric bimanual operations in the bimanual taxonomy literature \cite{Franziska2022Bimanual}. For the bimanual system, Edsinger and Kemp \cite{edsinger2007two} emphasized the importance of \textquotedblleft task-relevant features\textquotedblright\ in its design. Specifically, task-relevant features refer to key factors closely related to a specific task operation, which influence how the robot performs actions and collaborates effectively with humans, such as the shape and size of objects, task timing, and so on.

Tarbouriech et al. \cite{Sonny2019Admittance} considered the gravity and center of mass of the object as task-relevant features in their study of human-robot collaborative manipulation, ensuring that the object being transported remains balanced. Krebs and Asfour \cite{Franziska2022Bimanual} considered hand posture and velocity, as well as the contact surface between the sponge and the bowl, as task-relevant features in their study of bimanual dishwashing, to collect data for bimanual dishwashing tasks. Recently, Zhu et al. \cite{Jihong2024Do} proposed a bimanual interactive dressing strategy. The end-effector position of the interactive robot is used as the task-relevant feature, which is then transmitted to the dressing robot. This information is utilized to compute the human arm posture and design the dressing trajectories for the dressing robot. In this paper, the design of the spherical dressing coordinate system not only facilitates encoding but also provides the task-relevant features for bimanual assistance, namely the azimuthal angles of the spherical coordinate system.

Others have also studied bimanual assisted dressing tasks, such as those by Tamei et al. \cite{Tomoya2011Reinforcement}, Joshi et al. \cite{joshi2019framework}, and Clegg et al. \cite{Alexander2020Learning}. However, most of these studies focus on a \textit{one-robot-to-one-arm} setup, where each robot dresses a single human arm. In contrast, the approach described in this paper involves a \textit{two-robot-to-one-arm} setup. A similar study by Zhu et al. \cite{Jihong2024Do} also addresses \textit{two-robot-to-one-arm} systems, but their research is focused on asymmetrical bimanual manipulation, which is fundamentally different from that described here.

Additionally, it is important to note that although this paper focuses on the "bimanual dressing strategy for tight-fitting clothing," the main emphasis is on the study of dressing trajectories. Therefore, the paper does not provide an in-depth analysis of the material properties of the clothing or the mechanical aspects of the dressing process.

\section{METHOD}
\label{sec:method}

In this section, we introduce a spherical dressing coordinate system to facilitate encoding of bimanual dressing strategy. Then we apply the GMM/GMR imitation learning to bimanual dressing.

Before proposing the specific methods, the following assumptions are made:
\begin{enumerate}[label=\arabic*)]
  \item The human arm remains stationary.
  \item The clothing is already placed over the sleeve cuff, and the end-effector has grasped the cuff.
  \item Only one human arm is being dressed.
\end{enumerate}

\subsection{Spherical Dressing Coordinate}
\label{subsec:method:spher_dress_coord}

\begin{figure}[htbp]
  \centering %
  \includegraphics[width=0.45\textwidth]{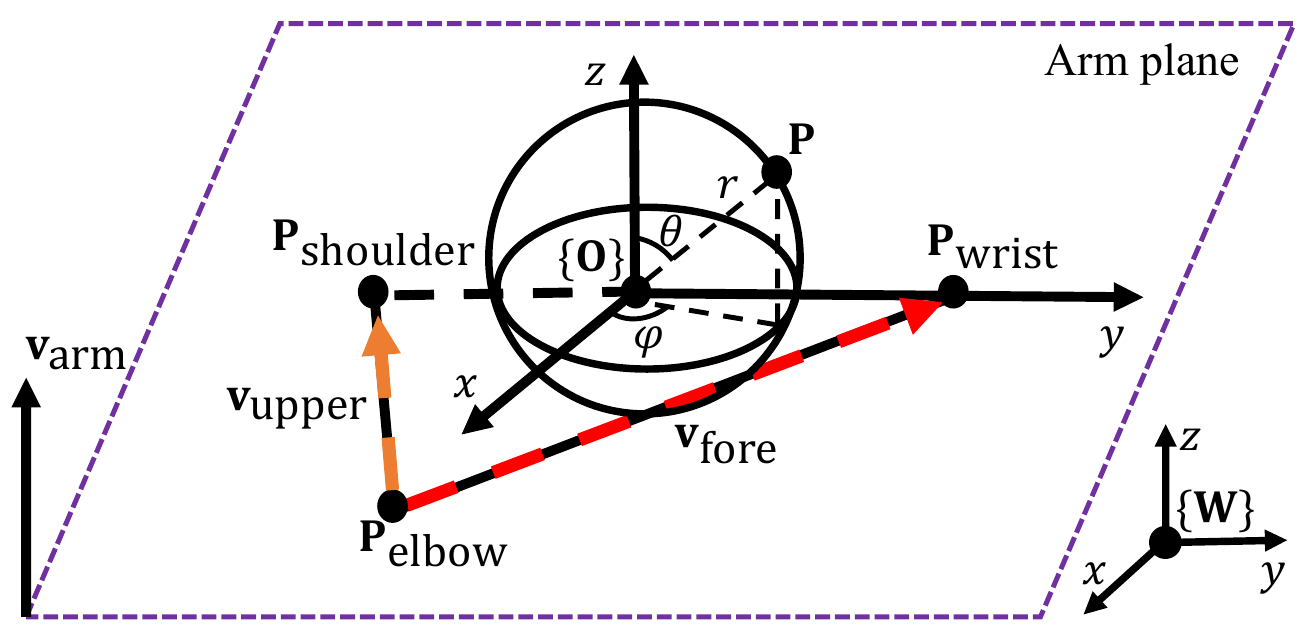} %
  \caption{Spherical coordinate system. The center of the sphere, \( \mathbf{O} \), is the foot of the perpendicular from point \( \mathbf{P}_{\text{elbow}} \) to line segment \( \overline{\mathbf{P}_{\text{wrist}} \mathbf{P}_{\text{shoulder}}} \). The x-axis is defined as the axis pointing from the sphere center \( \mathbf{O} \) to \( \mathbf{P}_{\text{elbow}} \), while the z-axis is aligned with the direction of \( \mathbf{v}_{\text{arm}} \). Any point \( \mathbf{P}(x, y, z) \) in the Cartesian coordinate system centered at \( \mathbf{O} \) can be converted into spherical coordinates \( (r, \theta, \varphi) \).} %
  \label{fig:spheric_coord:coord} %
   \vspace{-3mm}
\end{figure}

As shown in Fig. \ref{fig:spheric_coord:coord}, the world coordinate system is defined as \(\{\mathbf{W}\}\), where the three joints of the human arm (wrist, elbow, and shoulder) are simplified into three points, represented by \( \mathbf{P}_{\text{wrist}} \), \( \mathbf{P}_{\text{elbow}} \), and \( \mathbf{P}_{\text{shoulder}} \), respectively. The forearm and upper arm of the human arm are simplified into two vectors, represented by \( \mathbf{v}_{\text{fore}} \) and \( \mathbf{v}_{\text{upper}} \), respectively, where
\begin{equation}
\label{eq:spher_dress_coord:v_forearm_define}
\mathbf{v}_{\text{fore}} = \mathbf{P}_{\text{wrist}} - \mathbf{P}_{\text{elbow}}
\end{equation}
\begin{equation}
\label{eq:spher_dress_coord:v_upperarm_define}
\mathbf{v}_{\text{upper}} = \mathbf{P}_{\text{shoulder}} - \mathbf{P}_{\text{elbow}}
\end{equation}

The plane defined by the points \( \mathbf{P}_{\text{wrist}} \), \( \mathbf{P}_{\text{elbow}} \), and \( \mathbf{P}_{\text{shoulder}} \) is referred to as the \textquotedblleft arm plane\textquotedblright. The normal vector to the arm plane is denoted as \( \mathbf{v}_{\text{arm}} \), which is determined by the cross product of \( \mathbf{v}_{\text{fore}} \) and \( \mathbf{v}_{\text{upper}} \), i.e.
\begin{equation}
\label{eq:spher_dress_coord:v_arm_define}
\mathbf{v}_{\text{arm}} = \mathbf{v}_{\text{fore}} \times \mathbf{v}_{\text{upper}}
\end{equation}

The angle between \( \mathbf{v}_{\text{fore}} \) and \( \mathbf{v}_{\text{upper}} \) is defined as the elbow joint angle, denoted as \( \psi \), i.e.
\begin{equation}
\label{eq:spher_dress_coord:elbow_angle_define}
\psi = \cos^{-1} \left( \frac{\mathbf{v}_{\text{fore}} \cdot \mathbf{v}_{\text{upper}}}{|\mathbf{v}_{\text{fore}}| |\mathbf{v}_{\text{upper}}|} \right)
\end{equation}

To define the spherical coordinate system, it is necessary to first determine its origin. Let \( \mathbf{O} \) be the foot of the perpendicular from point \( \mathbf{P}_{\text{elbow}} \) to line segment \(\overline{\mathbf{P}_{\text{wrist}} \mathbf{P}_{\text{shoulder}}}\), and take \( \mathbf{O} \) as the origin of the spherical coordinate system.

Next, it is necessary to define the Cartesian coordinate system \(\{\mathbf{O}\}\) fixed at the origin of the spherical coordinate system. The \(x\)-axis of coordinate system \(\{\mathbf{O}\}\) is defined as the axis pointing in the direction of the vector from \( \mathbf{O} \) to \( \mathbf{P}_{\text{elbow}} \). The \(z\)-axis of coordinate system \(\{\mathbf{O}\}\) is defined as the axis pointing in the direction of \( \mathbf{v}_{\text{arm}} \). The \(y\)-axis of coordinate system \(\{\mathbf{O}\}\) is determined by the right-hand rule.

Based on the Cartesian coordinate system \( \{\mathbf{O}\} \), the spherical coordinate system can be defined. For any point \( \mathbf{P} \) in space, if the coordinates of \( \mathbf{P} \) in the Cartesian coordinate system \( \{\mathbf{O}\} \) are \( (x, y, z) \), then the coordinates of \( \mathbf{P} \) in the spherical coordinate system \( (r, \theta, \varphi) \) are determined by \eqref{eq:spher_dress_coord:r_theta_phi_define}.

\begin{equation}
\label{eq:spher_dress_coord:r_theta_phi_define}
\begin{aligned}
    r &= \sqrt{x^2 + y^2 + z^2}, \quad r \geq 0, \\
    \theta &= \arccos\left(\frac{z}{r}\right), \quad 0 \leq \theta \leq \pi, \\
    \varphi &= \arctan\left(\frac{y}{x}\right), \quad -\pi \leq \varphi < \pi.
\end{aligned}
\end{equation}
where, \( r \) is the radial distance, representing the distance between point \( \mathbf{P} \) and the origin \( \mathbf{O} \). \( \theta \) is the polar angle, representing the angle between the \( z \)-axis and the radial line segment \( \mathbf{O}\mathbf{P} \). \( \varphi \) is the azimuthal angle, representing the angle between the \( x \)-axis and the orthogonal projection of the radial line segment \( \mathbf{O}\mathbf{P} \) on the \( xy \)-plane (i.e., arm plane). The range of \( \varphi \) is defined from \( -\pi \) to \( \pi \) for convenience in subsequent calculations.

\begin{figure}[tbp]
  \centering %
  \includegraphics[width=0.45\textwidth]{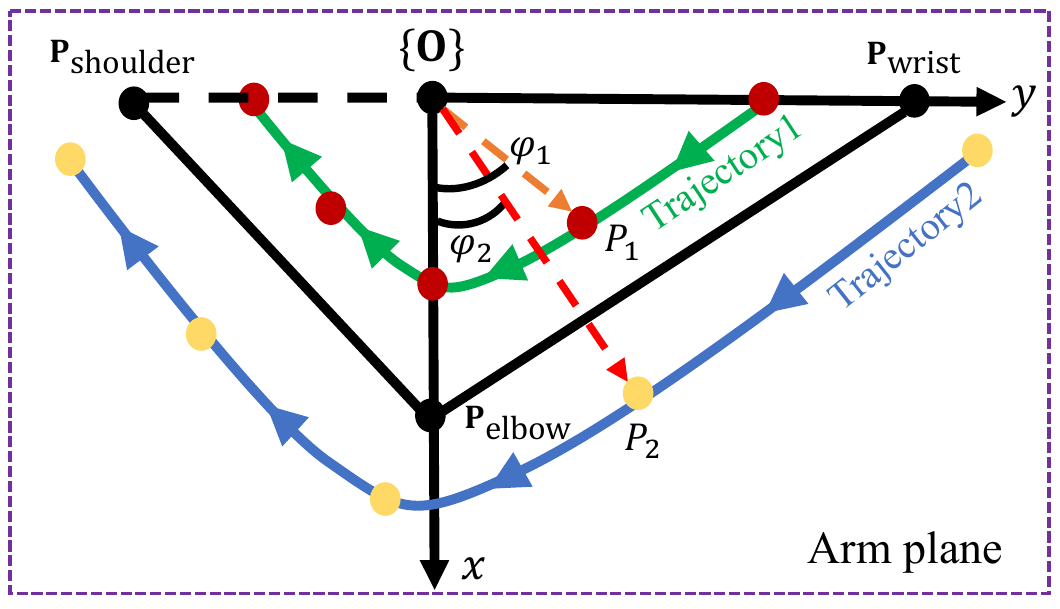} %
  \caption{Schematic of the arm plane and dual-arm trajectories: Arm 1 (\( \mathbf{P}_1 \), red) follows trajectory 1 (green) from wrist to shoulder, and arm 2 (\( \mathbf{P}_2 \), yellow) follows trajectory 2 (blue) along a similar path. The azimuthal angles \( \varphi_1 \) and \( \varphi_2 \) of the arms are correlated.} %
  \label{fig:spher_coord:traj} %
   \vspace{-8mm}
\end{figure}

Next, a qualitative analysis of the task-relevant features is performed between the two robotic arms, i.e., the azimuthal angle. First, the arm-plane is shown in Fig. \ref{fig:spher_coord:traj}, where the green and blue curves represent the projections of the first and second robotic arm trajectories onto the arm-plane, respectively. Red point \( \mathbf{P}_1 \) and yellow point \( \mathbf{P}_2 \) represent the positions of the end effectors of the two robotic arms at the same moment, and \( \varphi_1 \) and \( \varphi_2 \) are the corresponding azimuthal angles. During the human operation of bimanual dressing, the two trajectories follow similar paths traced by the human arms, i.e., they both first move along the forearm to the elbow joint, then turn at the elbow joint, and continue along the upper arm to the shoulder joint. Correspondingly, \( \varphi_1 \) and \( \varphi_2 \) decrease from approximately \( \pi/2 \) (at the wrist joint) to approximately 0 (at the elbow joint) and finally stop together at approximately \( -\pi/2 \) (at the shoulder joint). This suggests that there is a certain correlation between \( \varphi_1 \) and \( \varphi_2 \) in the spherical coordinate system. Therefore,  \( \varphi_1 \) and \( \varphi_2 \) can be considered as the task-relevant features for bimanual operation. In Section \ref{sec:experiment}, a data-driven quantitative analysis of their relationship is conducted.

Compared to the dressing coordinate system established in \cite{Jihong2024Do}, the spherical coordinate system proposed in this paper has the advantage of simpler coordinate transformation relationships, without the need for complex parameter settings.

\subsection{Bimanual Dressing Policy in Dressing Coordinate}
\label{subsec:method:dual_arms_lfd}

For learning from demonstration (LfD) in the spherical dressing coordinate system,  GMM and GMR are utlised to learn and generate the dressing policy \cite{Sylvain2007Learning}. The imitation learning approach based on GMM/GMR enables fast movement retrieval and flexible input-output arrangements, making it well-suited for our task. A brief overview of the principles of GMM/GMR-based LfD is presented.

GMM models the joint distribution of the demonstration data. The demonstrated data is denoted $\boldsymbol{\xi}$.
\begin{equation}
\label{eq:xi_define}
\boldsymbol{\xi} = \begin{bmatrix}{\boldsymbol{\xi}}^\mathcal{I},  {\boldsymbol{\xi}}^\mathcal{O}\end{bmatrix}^{T}
\end{equation}
with ${\boldsymbol{\xi}}^\mathcal{I}$ is the input and ${\boldsymbol{\xi}}^\mathcal{O}$ the output. The probability density function $p({\boldsymbol{\xi}}_{i})$ is estimated with $K$ Gaussian distributions, i.e.
\begin{equation}
\label{eq:p_xi_define}
p({\boldsymbol{\xi} }_{i})=\sum_{k=1}^{K}\pi_{k}\mathcal{N}({\boldsymbol{\xi}}_{i}\big|  {\boldsymbol{\mu}}_{k},\boldsymbol{\Sigma}_{k})
\end{equation}
where $\boldsymbol{\mu}_k$ and $\boldsymbol{\Sigma} _{k}$ are mean and variance of the $k$th Gaussian, respectively. To determine an optimal number of Gaussian $K$ for GMM, the Bayesian information criterion is used \cite{Schwarz1978Estimating} for balancing the model complexity and representation quality.

Once $K$ is selected, the GMM with K-means clustering is utalised and then employ the expectation–maximization (EM) algorithm to iteratively compute the model parameters $\left\{\pi _{k},\boldsymbol{\mu}_k,\boldsymbol{\Sigma}_{k}\right\}^K_{k=1}$.

The resulting $K$ Gaussian parameters can be decomposed as
\begin{equation}
\label{eq:dual_lfd:mu_k_and_sigma_k_define}
\boldsymbol{\mu }_{k} = \begin{bmatrix}\boldsymbol{\mu }_{k}^\mathcal {I} \\ \boldsymbol{\mu }_{k}^\mathcal {O} \end{bmatrix}, \quad  
\boldsymbol{\Sigma }_{k} = \begin{bmatrix}\boldsymbol{\Sigma }_{k}^\mathcal {I} & \hat{ \boldsymbol{\Sigma }}_{k}^{\mathcal {I}\mathcal {O}} \\  \boldsymbol{\Sigma }_{k}^{\mathcal {O}\mathcal {I}} & \hat{ \boldsymbol{\Sigma }}_{k}^\mathcal {O} \end{bmatrix}
\end{equation}

GMR utilizes the parameters obtained from the GMM to perform regression by computing the conditional distribution of the output $\boldsymbol{\xi}^\mathcal{O}$ given the input $\boldsymbol{\xi}^\mathcal{I}$ \cite{Manuel2009TaskLevel}. For each Gaussian component $k$, the conditional distribution is defined as
\[
\mathcal{N}\left(\boldsymbol{\xi}^\mathcal{O} \big| \boldsymbol{\mu}_k^{\mathcal{O}|\mathcal{I}}, \boldsymbol{\Sigma}_k^{\mathcal{O}|\mathcal{I}} \right),
\]
where the conditional mean and covariance are computed by
\begin{equation}
\label{eq:dual_lfd:mu_cond_and_Sigma_cond_define}
\begin{aligned}
\boldsymbol{\mu}_k^{\mathcal{O}|\mathcal{I}} &= \boldsymbol{\mu}_k^\mathcal{O} + \hat{\boldsymbol{\Sigma}}_k^{\mathcal{O}\mathcal{I}} \left( \boldsymbol{\Sigma}_k^\mathcal{I} \right)^{-1} \left( \boldsymbol{\xi}^\mathcal{I} - \boldsymbol{\mu}_k^\mathcal{I} \right), \\
\boldsymbol{\Sigma}_k^{\mathcal{O}|\mathcal{I}} &= \hat{\boldsymbol{\Sigma}}_k^\mathcal{O} - \hat{\boldsymbol{\Sigma}}_k^{\mathcal{O}\mathcal{I}} \left( \boldsymbol{\Sigma}_k^\mathcal{I} \right)^{-1} \hat{\boldsymbol{\Sigma}}_k^{\mathcal{I}\mathcal{O}}.
\end{aligned}
\end{equation}

The posterior probability of each component given an input $\boldsymbol{\xi}^\mathcal{I}$ is then computed using Bayes' rule:
\begin{equation}
\label{eq:beta_define}
\beta_k(\boldsymbol{\xi}^\mathcal{I}) = \frac{\pi_k \, \mathcal{N}\left( \boldsymbol{\xi}^\mathcal{I} \big| \boldsymbol{\mu}_k^\mathcal{I}, \boldsymbol{\Sigma}_k^\mathcal{I} \right)}{\sum_{j=1}^{K} \pi_j \, \mathcal{N}\left( \boldsymbol{\xi}^\mathcal{I} \big| \boldsymbol{\mu}_j^\mathcal{I}, \boldsymbol{\Sigma}_j^\mathcal{I} \right)}.
\end{equation}

Finally, the overall regression output is obtained by taking the weighted sum of the conditional means:
\begin{equation}
\label{eq:mu_output_define}
\boldsymbol{\mu}^\mathcal{O}(\boldsymbol{\xi}^\mathcal{I}) = \sum_{k=1}^{K} \beta_k(\boldsymbol{\xi}^\mathcal{I}) \, \boldsymbol{\mu}_k^{\mathcal{O}|\mathcal{I}}.
\end{equation}

This GMR framework provides a smooth regression function that not only estimates the expected output but also quantifies the uncertainty associated with the predictions.

In this study, a static human arm posture and record the human arm posture are demonstrated and the bimanual dressing path, which is then converted into the spherical coordinate system. The trajectories of robotic arm 1 and robotic arm 2 in the spherical coordinate system are denoted as \( (r_1, \theta_1, \varphi_1) \) and \( (r_2, \theta_2, \varphi_2) \), respectively. The corresponding elbow joint angle of the human arm is denoted as \( \psi \).

\begin{figure}[tbp]
  \centering %
  \includegraphics[width=0.45\textwidth]{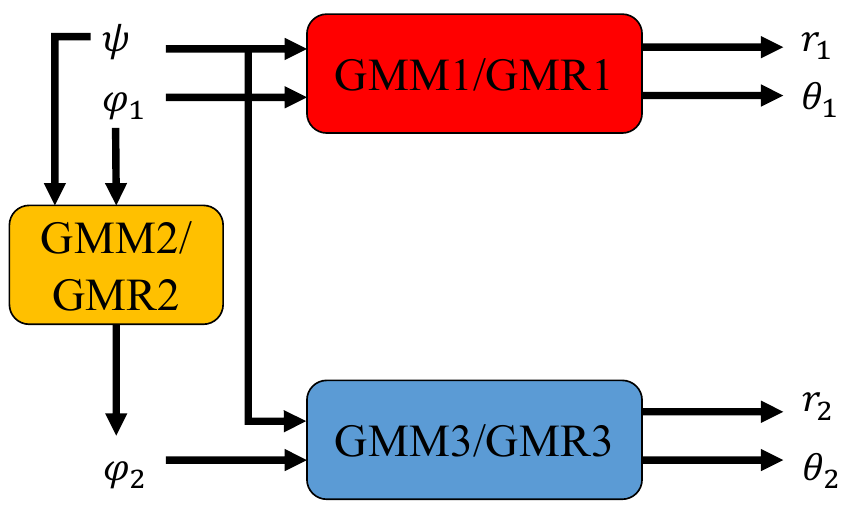} %
  \caption{Imitation learning for bimanual dressing using GMM/GMR: GMM1/GMR1 trains the first robotic arm's spherical coordinate trajectory, GMM3/GMR3 trains the second robotic arm's trajectory, and GMM2/GMR2 models the azimuthal angle relationship between the two robotic arms.} %
  \label{fig:dual_lfd:gmm} %
   \vspace{-3mm}
\end{figure}

The imitation learning strategy for bimanual dressing trajectories, used in this work is shown in the Fig. \ref{fig:dual_lfd:gmm}. In the training phase,  \( \psi \) and \( \varphi_1 \) are used as inputs and \( r_1 \) and \( \theta_1 \) as outputs to train the first GMM, thereby obtaining the relationship between \( \varphi_1 \) and \( r_1 \), \( \theta_1 \) under different elbow joint angles. Then,   \( \psi \) and \( \varphi_2 \) are used as inputs and \( r_2 \), \( \theta_2 \) as outputs to train the second GMM, thereby obtaining the relationship between \( \varphi_2 \) and \( r_2 \), \( \theta_2 \) under different elbow joint angles. Finally,  \( \psi \) and \( \varphi_1 \) are used as inputs and \( \varphi_2 \) as the output to train the third GMM model, thereby obtaining the relationship between \( \varphi_1 \) and \( \varphi_2 \) under different elbow joint angles. All demonstrations start from the wrist joint, pass through the elbow joint, and finally reach the shoulder joint. This hierarchical approach effectively decomposes the high-dimensional bimanual coordination problem into manageable components while preserving the essential coupling between arms. Such a strategy accommodates the non-linear relationships between joint angles and end-effector positions during dressing tasks, enabling smooth, synchronized movements that can adapt to the user's body configuration and potential movement limitations while maintaining appropriate tension across the garment.

In the testing phase, the human arm posture is captured and used to compute the corresponding elbow joint angle \(\psi\). Then, the input value of \(\varphi_1\) is set to range from \(\pi/2\) to \(-\pi/2\), and the corresponding \(r_1\) and \(\theta_1\) are generated using the first GMR. Next, \(\psi\) and \(\varphi_1\) are used as inputs to the third GMR to generate the corresponding \(\varphi_2\). Finally, \(\psi\) and \(\varphi_2\) are used as inputs to the second GMR to generate the corresponding \(r_2\) and \(\theta_2\). In this way, a bimanual dressing trajectory strategy is obtained for different human arm postures.

\section{EXPERIMENTS}
\label{sec:experiment}

\subsection{Data Collection}
\label{subsec:experiment:data_collect}
\begin{figure}[htbp]
  \centering  
  \begin{subfigure}[b]{0.23\textwidth}
    \includegraphics[width=\textwidth]{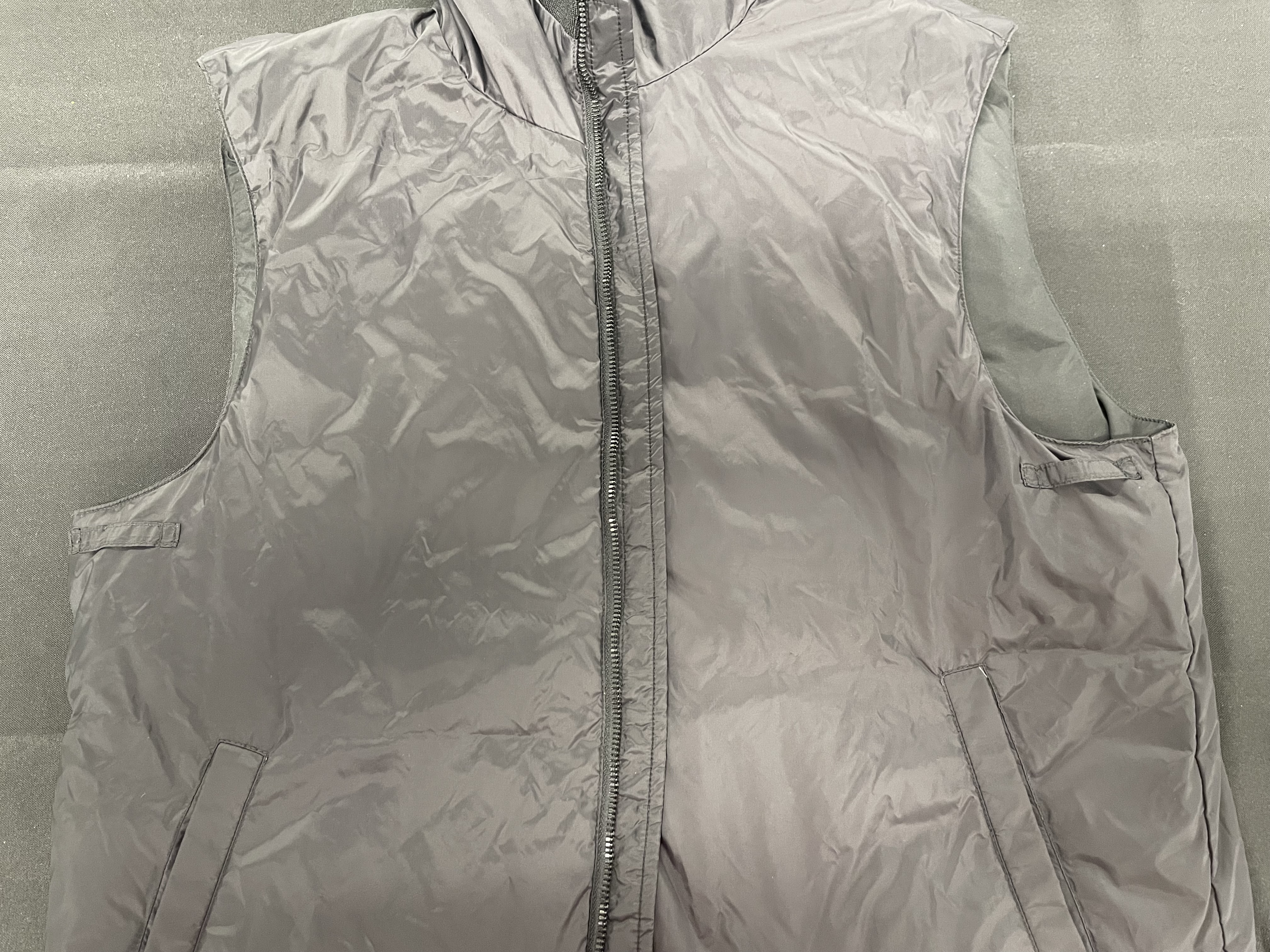}
    \caption{}
    \label{fig:data_collect:cloth_type:loose_cloth}
  \end{subfigure}
  \hfill 
  \begin{subfigure}[b]{0.23\textwidth}
    \includegraphics[width=\textwidth]{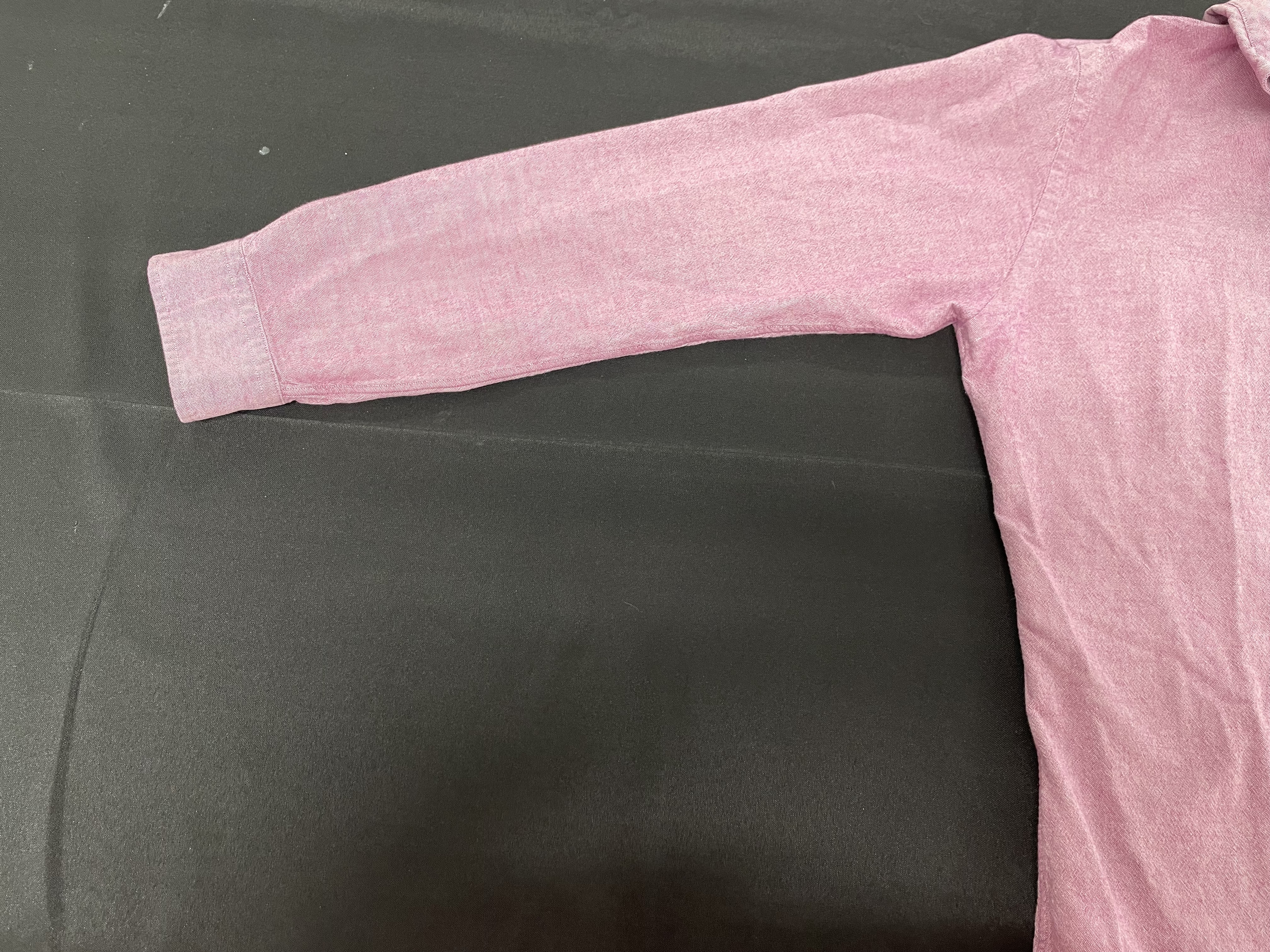}
    \caption{}
    \label{fig:data_collect:cloth_type:tight_cloth}
  \end{subfigure}
  \caption{Types of clothing used in the dressing experiment. (a) Loose sleeveless vest. (b) Tight long-sleeve shirt.}
  \label{fig:data_collect:cloth_type}
\end{figure}

Two types of clothing for the dressing experiment are used. As shown in the Fig. \ref{fig:data_collect:cloth_type}, the first type is a loose sleeveless vest, and the second type is a tight long-sleeve shirt. Based on the measurements, the armscye circumference of the loose sleeveless vest is \(55\) cm, and the armscye circumference of the long-sleeve shirt is \(45\) cm. Additionally, the arm circumference of the subject is also measured, which is \(30\) cm.

\begin{figure}[htbp]
  \centering
  \begin{subfigure}[b]{0.14\textwidth}
    \includegraphics[width=\textwidth]{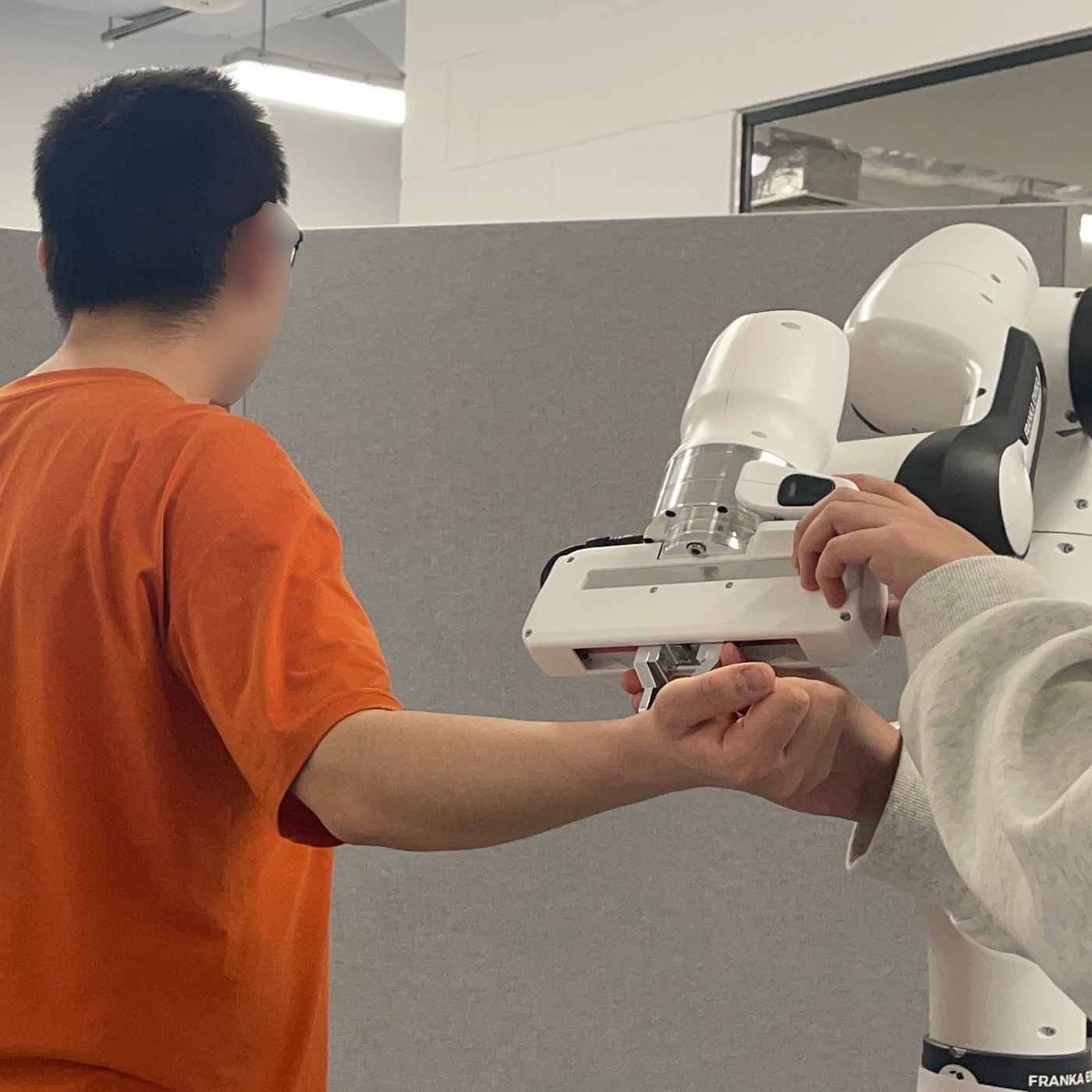}
    \caption{}
    \label{fig:data_collect:pose_traj:wrist_position}
  \end{subfigure}
  \hfill
  \begin{subfigure}[b]{0.14\textwidth}
    \includegraphics[width=\textwidth]{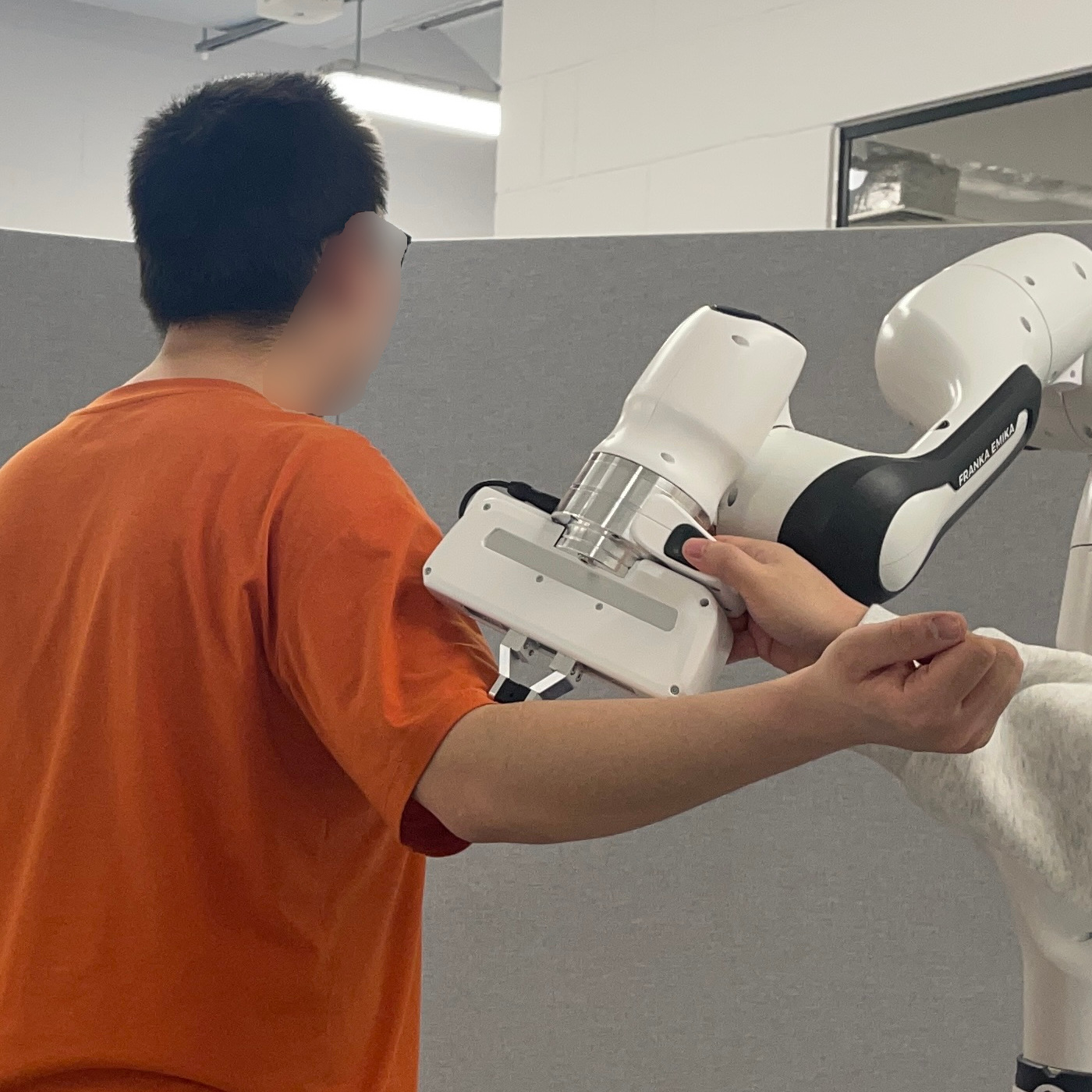}
    \caption{}
    \label{fig:data_collect:pose_traj:elbow_position}
  \end{subfigure}
  \hfill
  \begin{subfigure}[b]{0.14\textwidth}
    \includegraphics[width=\textwidth]{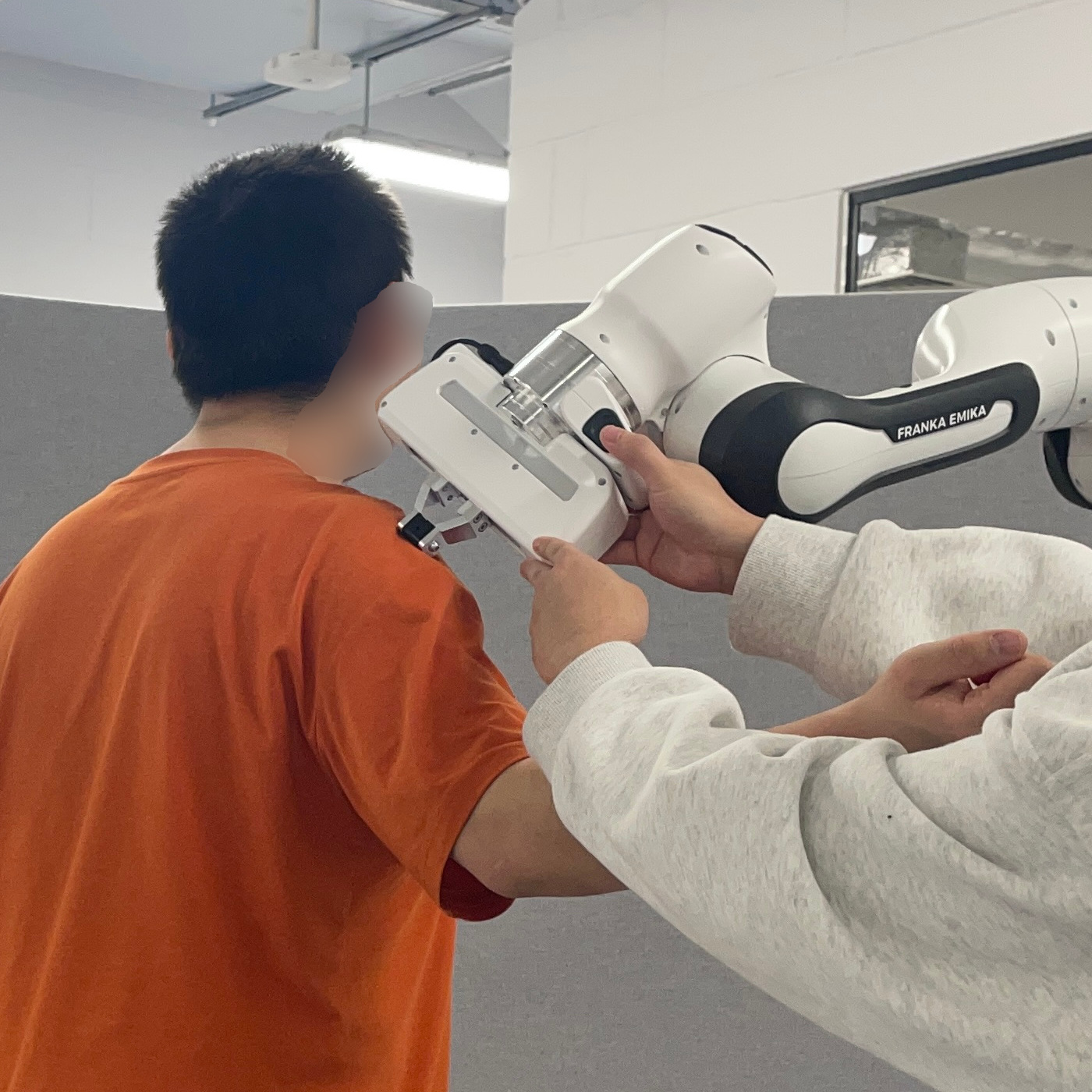}
    \caption{}
    \label{fig:data_collect:pose_traj:shoulder_position}
  \end{subfigure}

  \vskip\baselineskip
  
  \begin{subfigure}[b]{0.14\textwidth}
    \includegraphics[width=\textwidth]{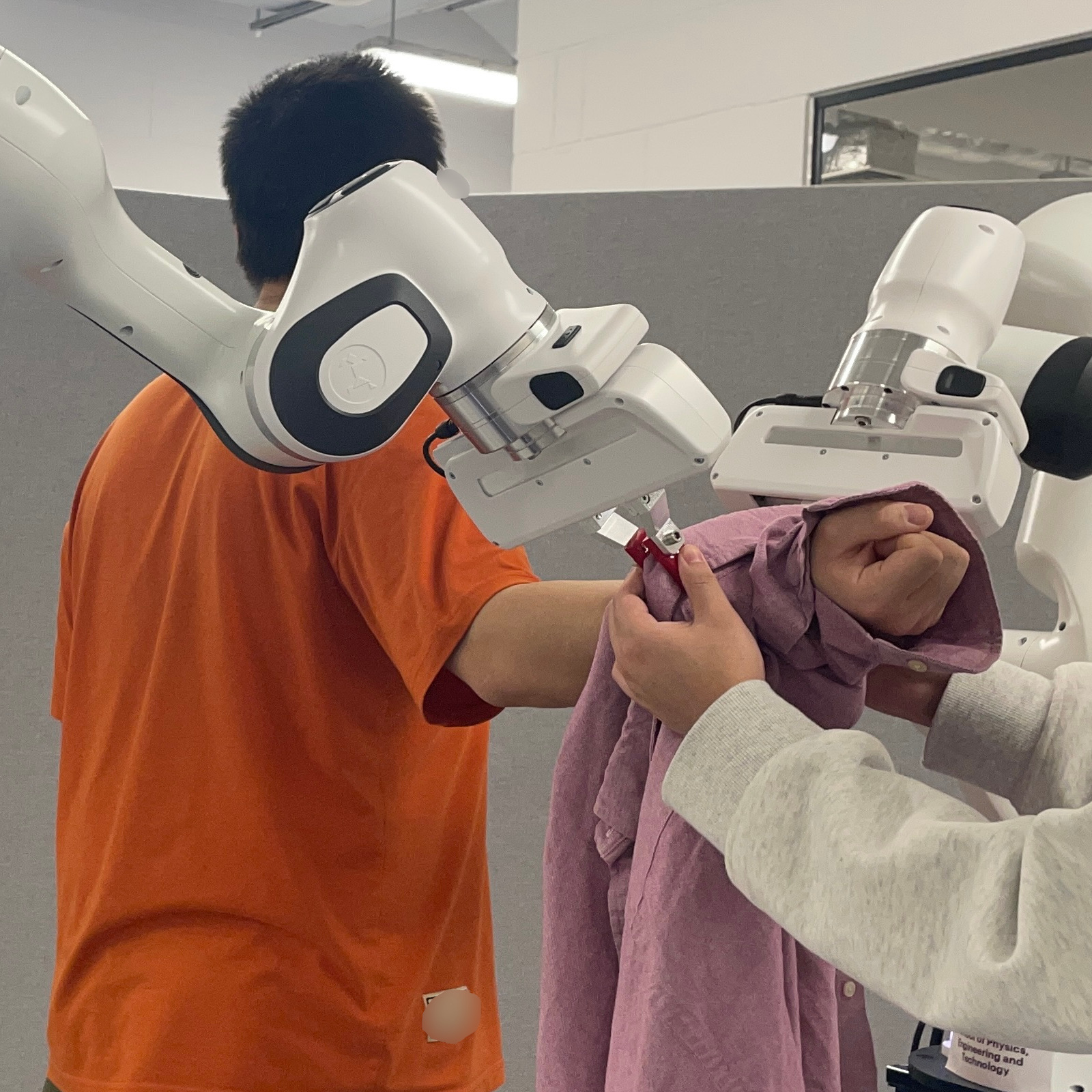}
    \caption{}
    \label{fig:data_collect:pose_traj:dressing_trajectory_1}
  \end{subfigure}
  \hfill
  \begin{subfigure}[b]{0.14\textwidth}
    \includegraphics[width=\textwidth]{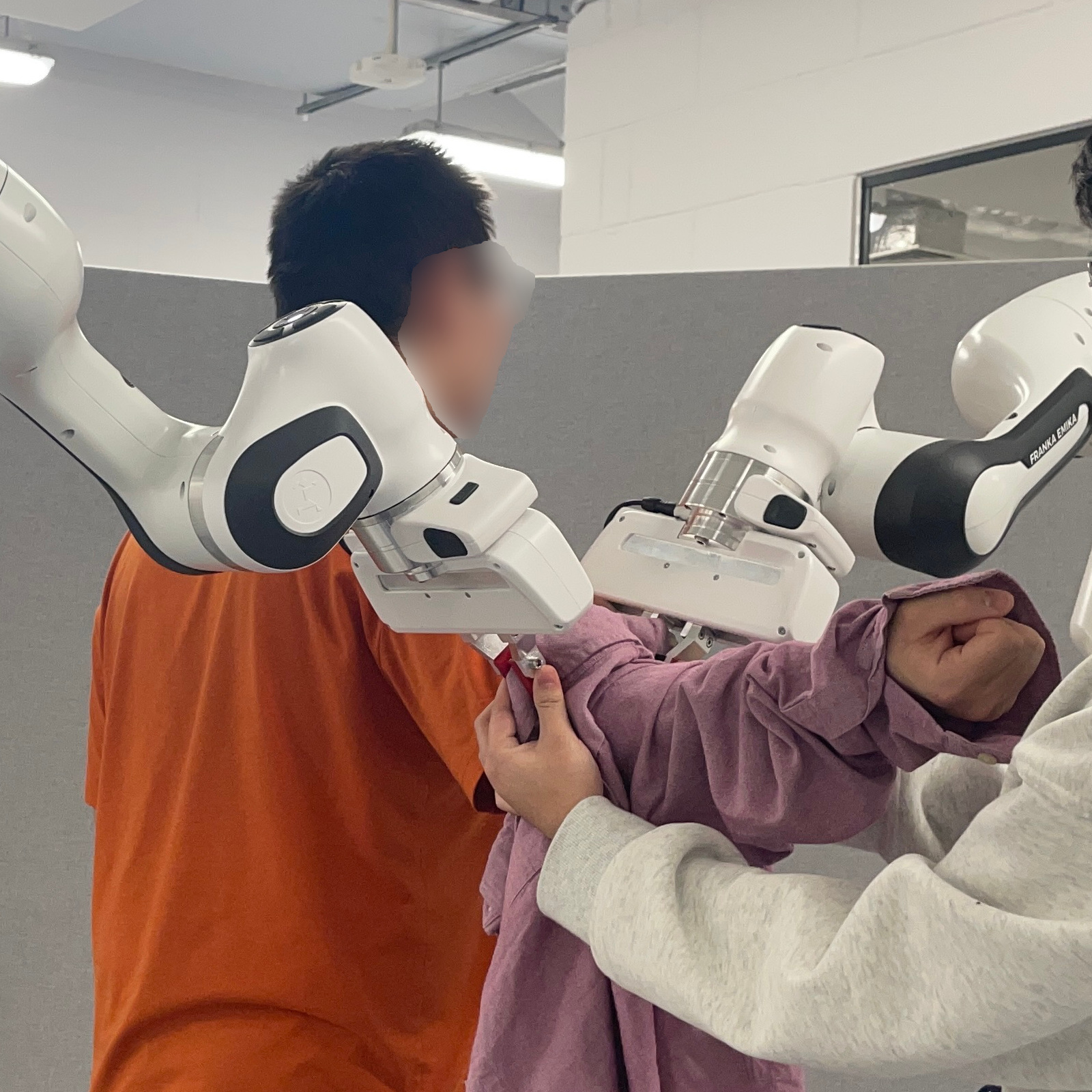}
    \caption{}
    \label{fig:data_collect:pose_traj:dressing_trajectory_2}
  \end{subfigure}
  \hfill
  \begin{subfigure}[b]{0.14\textwidth}
    \includegraphics[width=\textwidth]{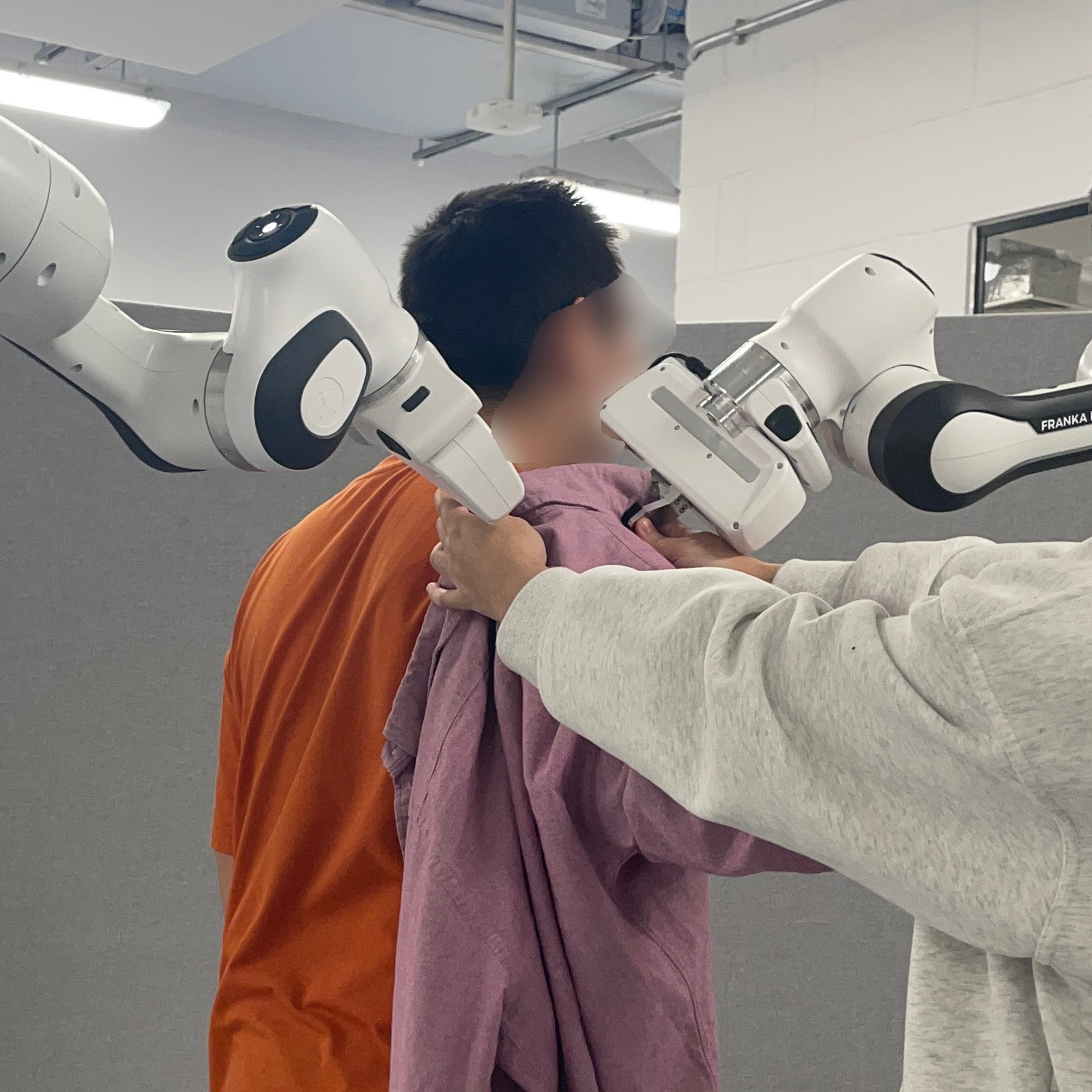}
    \caption{}
    \label{fig:data_collect:pose_traj:dressing_trajectory_3}
  \end{subfigure}

  \caption{Measurement process of the subject's arm joints and human-assisted dressing trajectories. (a)-(c) Positions of the wrist, elbow, and shoulder joints. (d)-(f) Human-assisted dressing trajectories.}
  \label{fig:data_collect:pose_traj}
\end{figure}

The human arm postures and corresponding dressing trajectories are collected. As shown in Fig. \ref{fig:data_collect:pose_traj}, the measurement process of the subject's wrist, elbow, and shoulder joints using the robotic arm's proprioceptive sensors is displayed in subfigures (a)-(c), while subfigures (d)-(f) show the process of collecting the human-assisted dressing trajectories. For tight garments, eight different tests were conducted with varying human arm postures, and the detailed process can be found in the video (see Supplementary material).

\subsection{Data Processing and Data Training}
\label{subsec:experiment:data_process}
\begin{figure*}[htbp]
  \centering
  \begin{minipage}[b]{0.9\textwidth}
    \centering
    \begin{subfigure}[b]{0.24\textwidth}
      \centering
      \includegraphics[width=\textwidth]{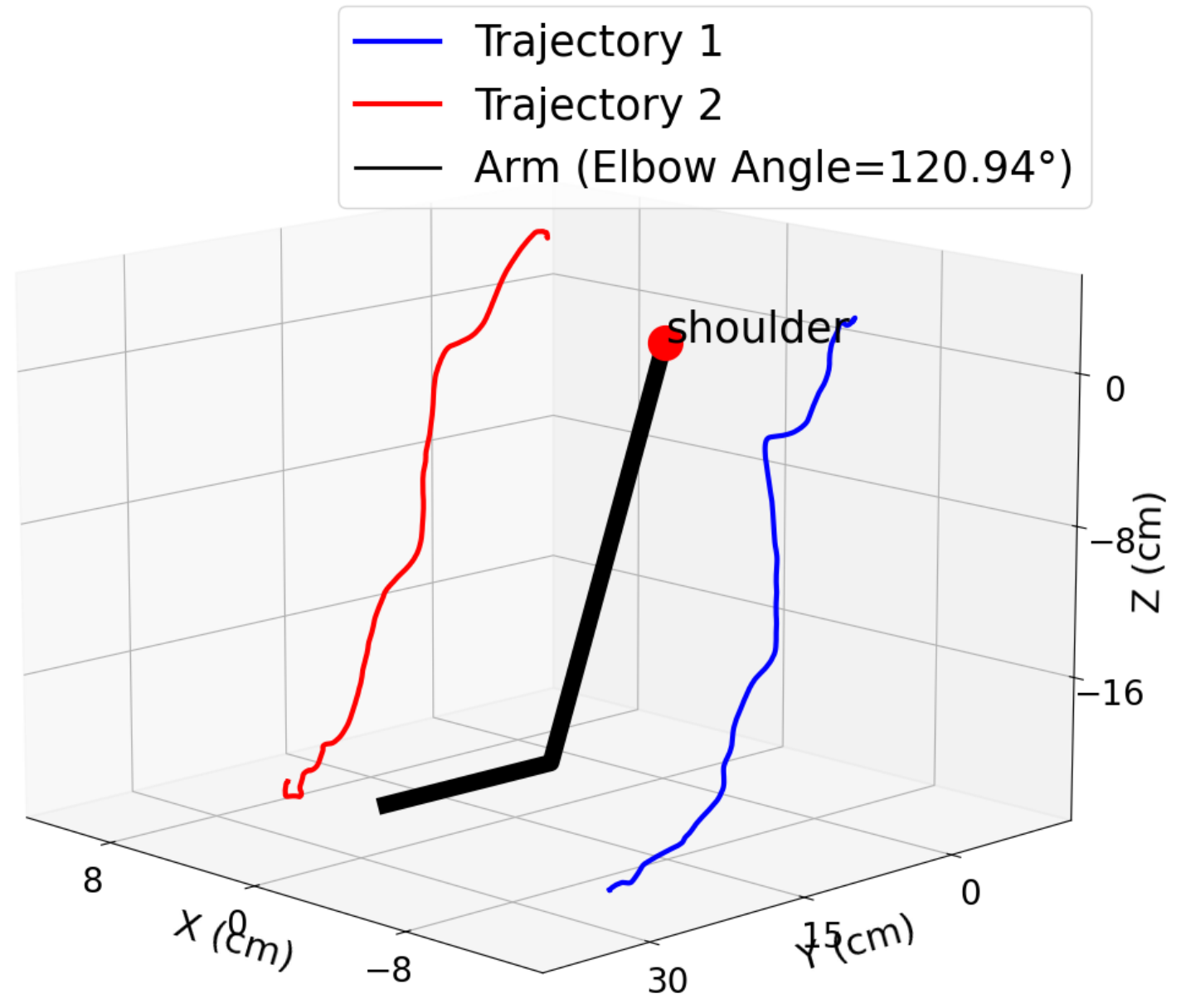}
      \caption{}
      \label{fig:data_process:pose_traj:traj_1}
    \end{subfigure}%
    \hfill
    \begin{subfigure}[b]{0.24\textwidth}
      \centering
      \includegraphics[width=\textwidth]{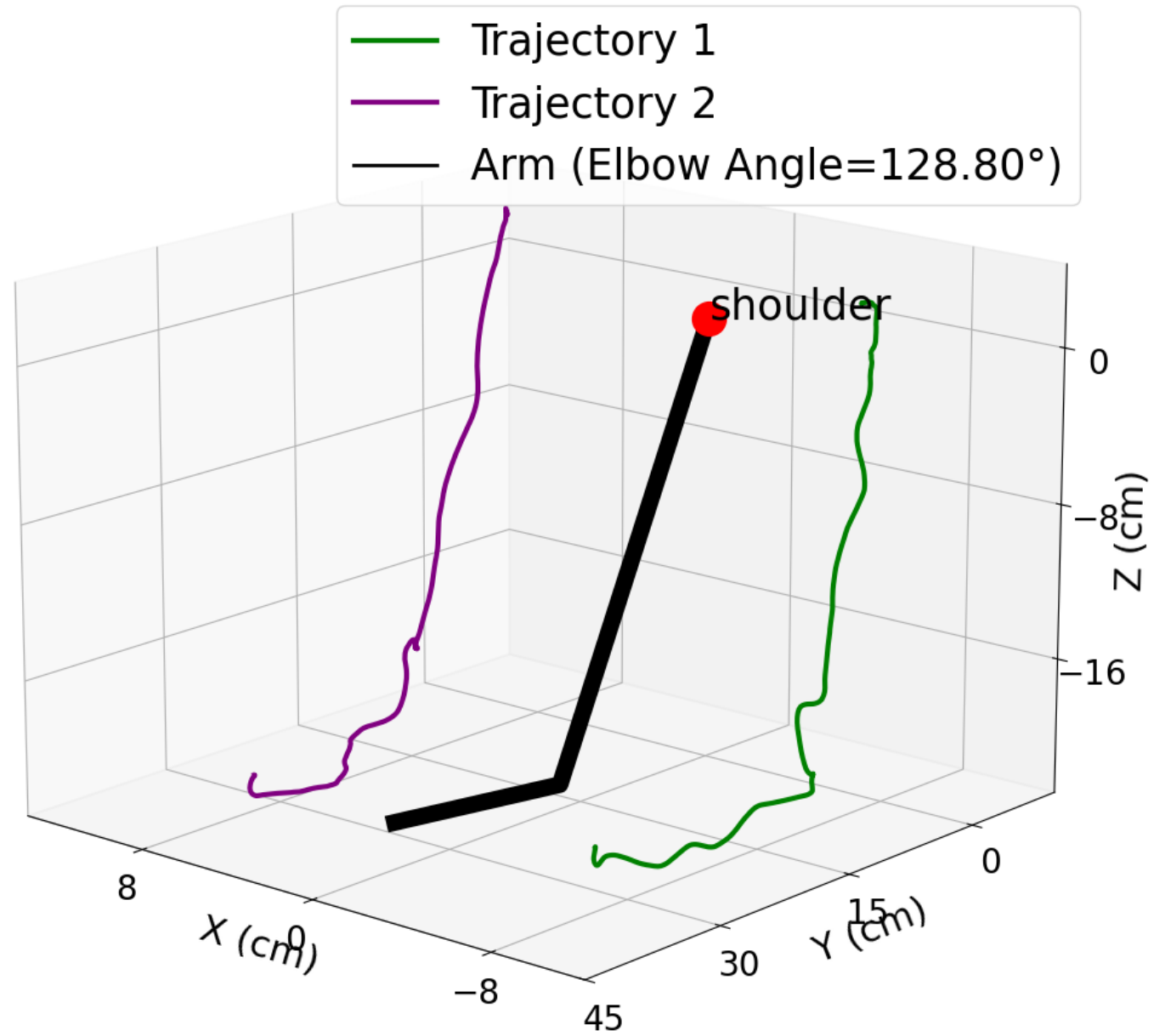}
      \caption{}
      \label{fig:data_process:pose_traj:traj_2}
    \end{subfigure}%
    \hfill
    \begin{subfigure}[b]{0.24\textwidth}
      \centering
      \includegraphics[width=\textwidth]{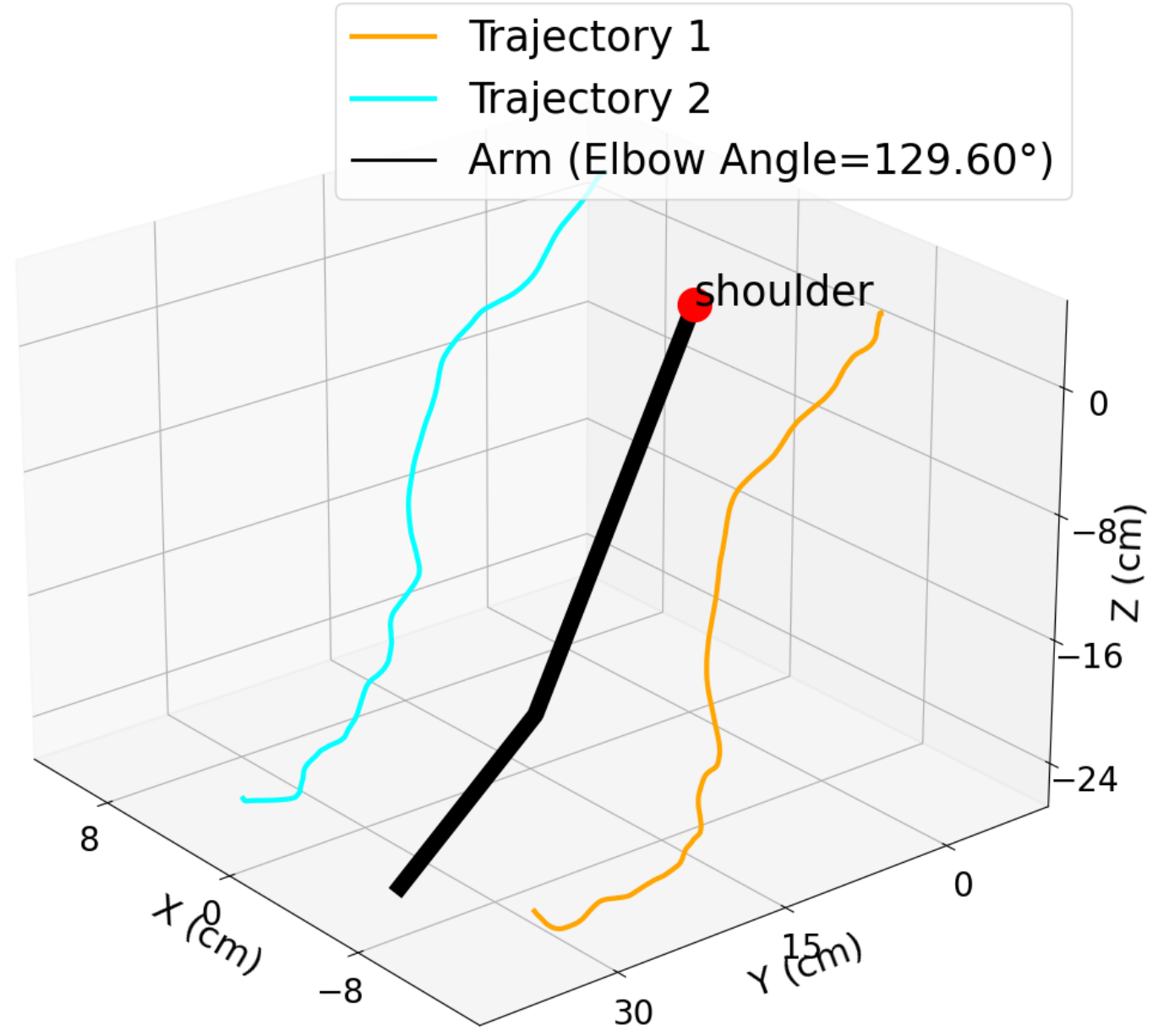}
      \caption{}
      \label{fig:data_process:pose_traj:traj_3}
    \end{subfigure}%
    \hfill
    \begin{subfigure}[b]{0.24\textwidth}
      \centering
      \includegraphics[width=\textwidth]{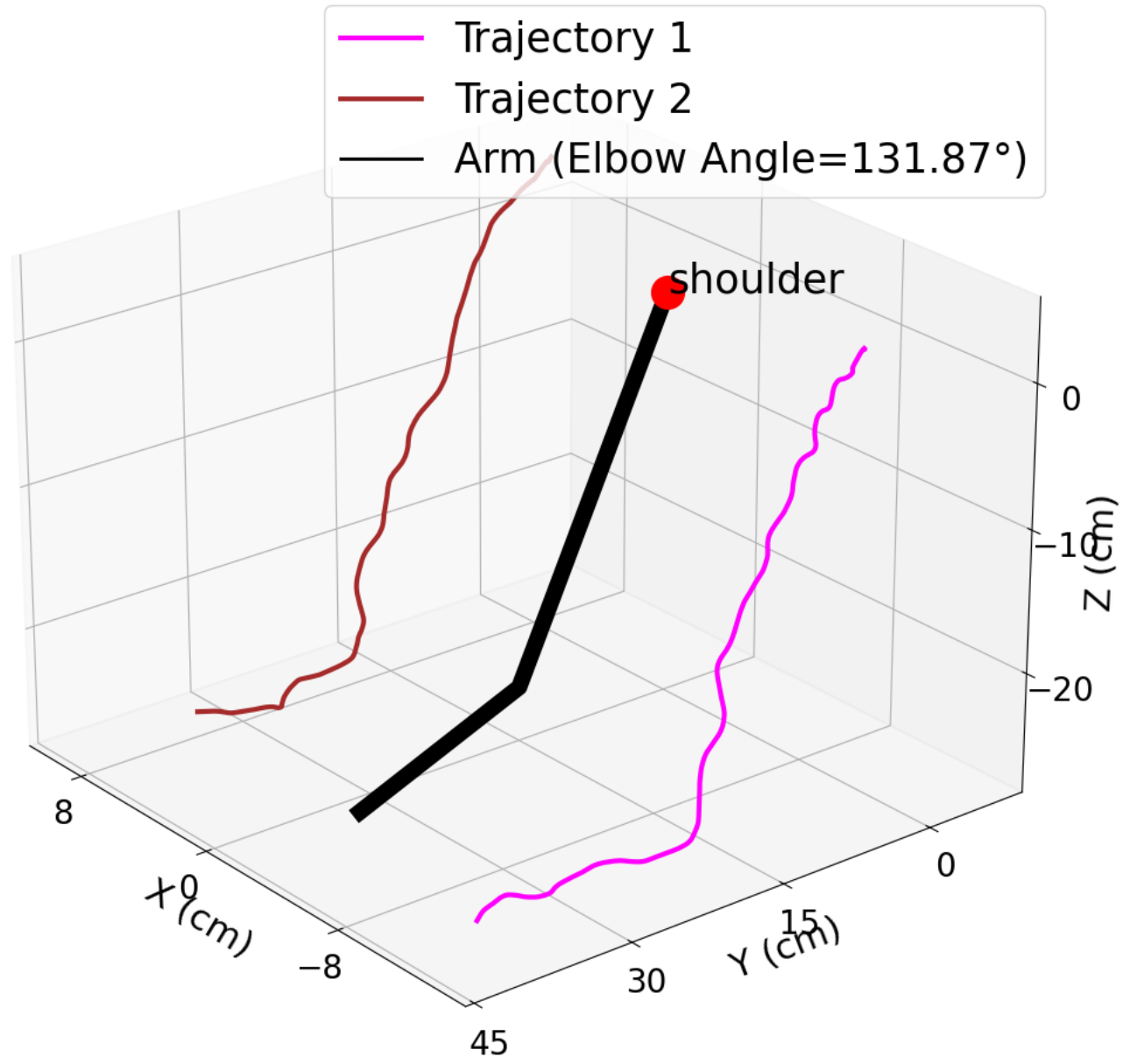}
      \caption{}
      \label{fig:data_process:pose_traj:traj_4}
    \end{subfigure}%

    \vskip\baselineskip

    \begin{subfigure}[b]{0.24\textwidth}
      \centering
      \includegraphics[width=\textwidth]{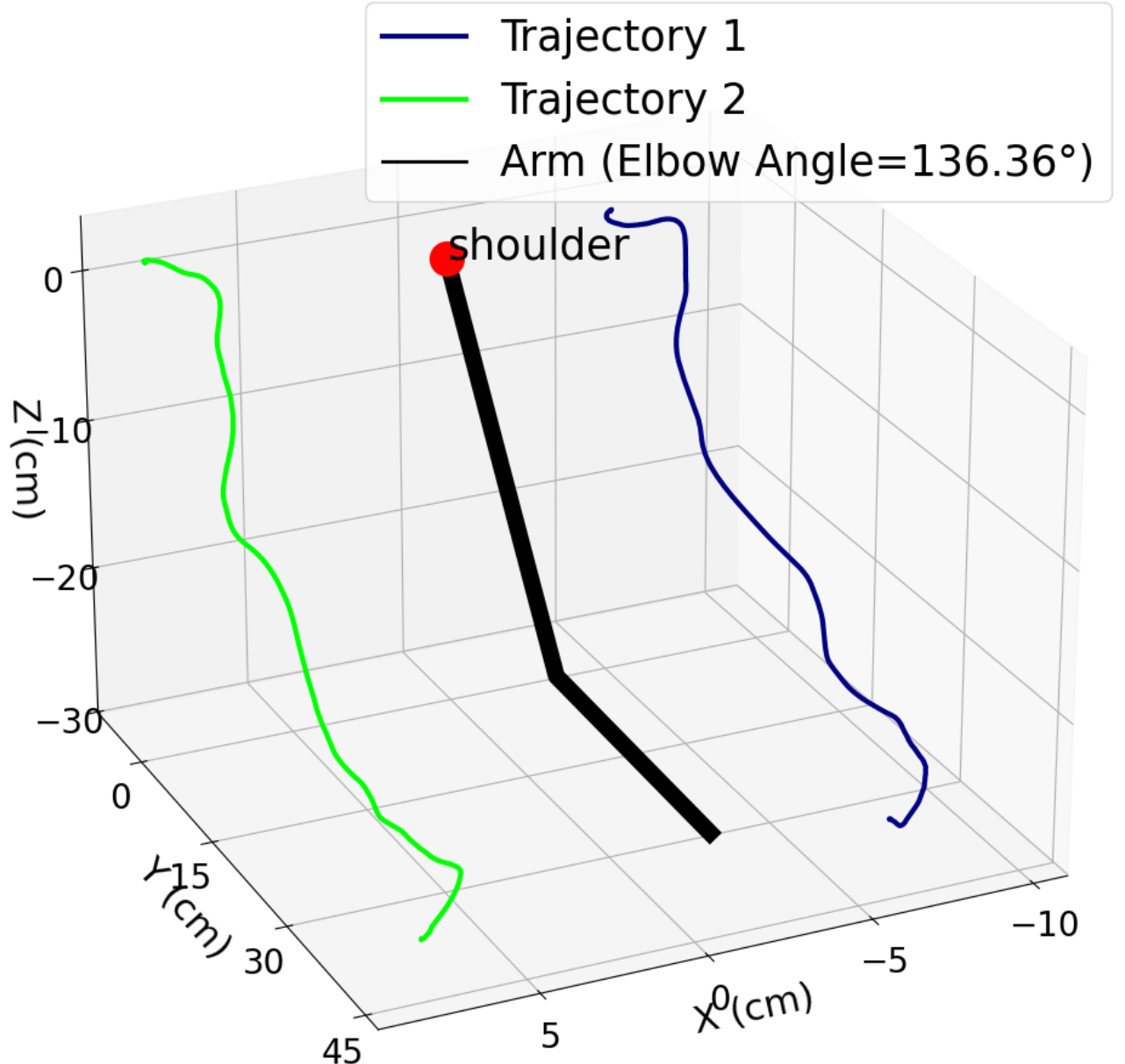}
      \caption{}
      \label{fig:data_process:pose_traj:traj_5}
    \end{subfigure}%
    \hfill
    \begin{subfigure}[b]{0.24\textwidth}
      \centering
      \includegraphics[width=\textwidth]{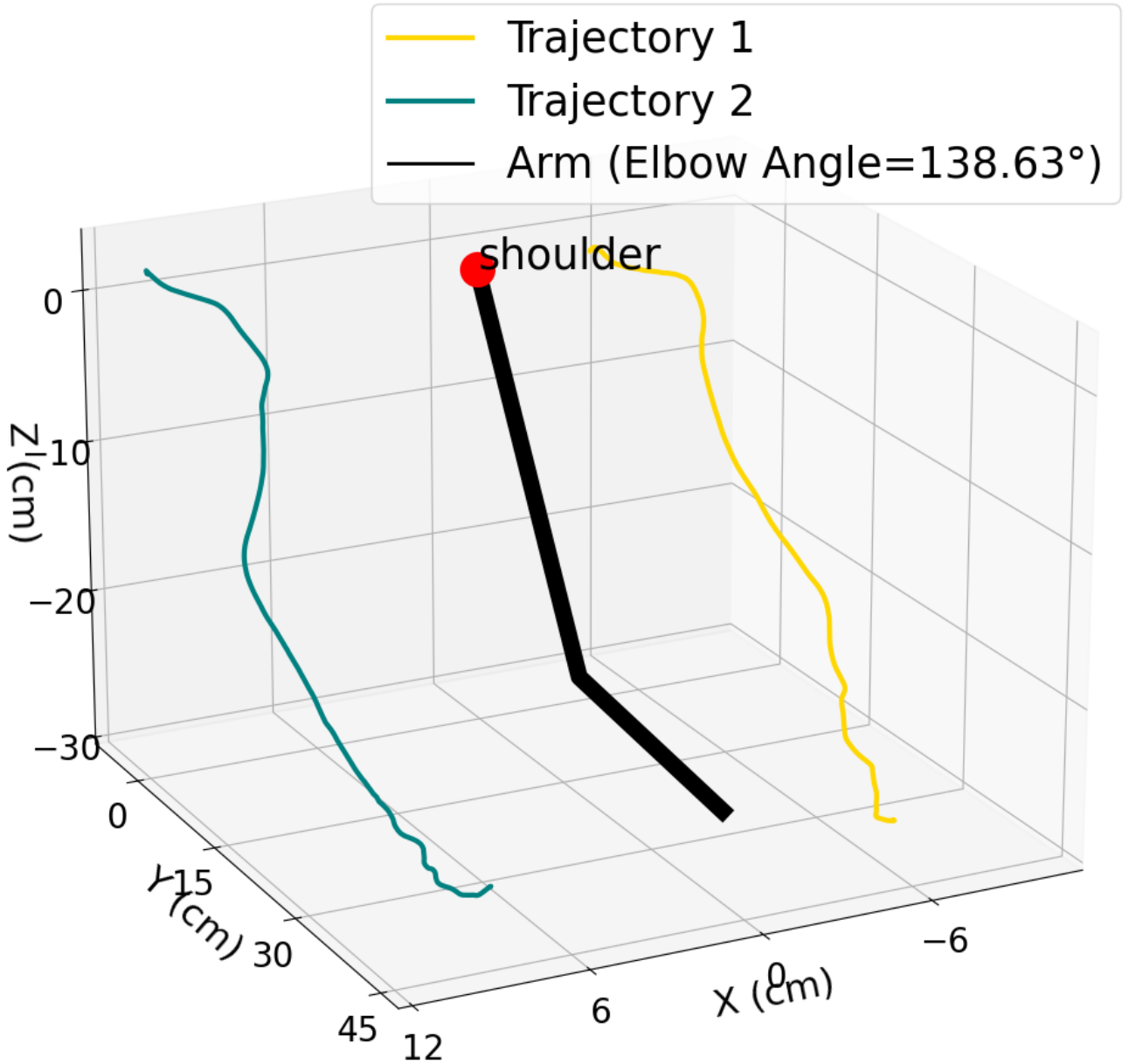}
      \caption{}
      \label{fig:data_process:pose_traj:traj_6}
    \end{subfigure}%
    \hfill
    \begin{subfigure}[b]{0.24\textwidth}
      \centering
      \includegraphics[width=\textwidth]{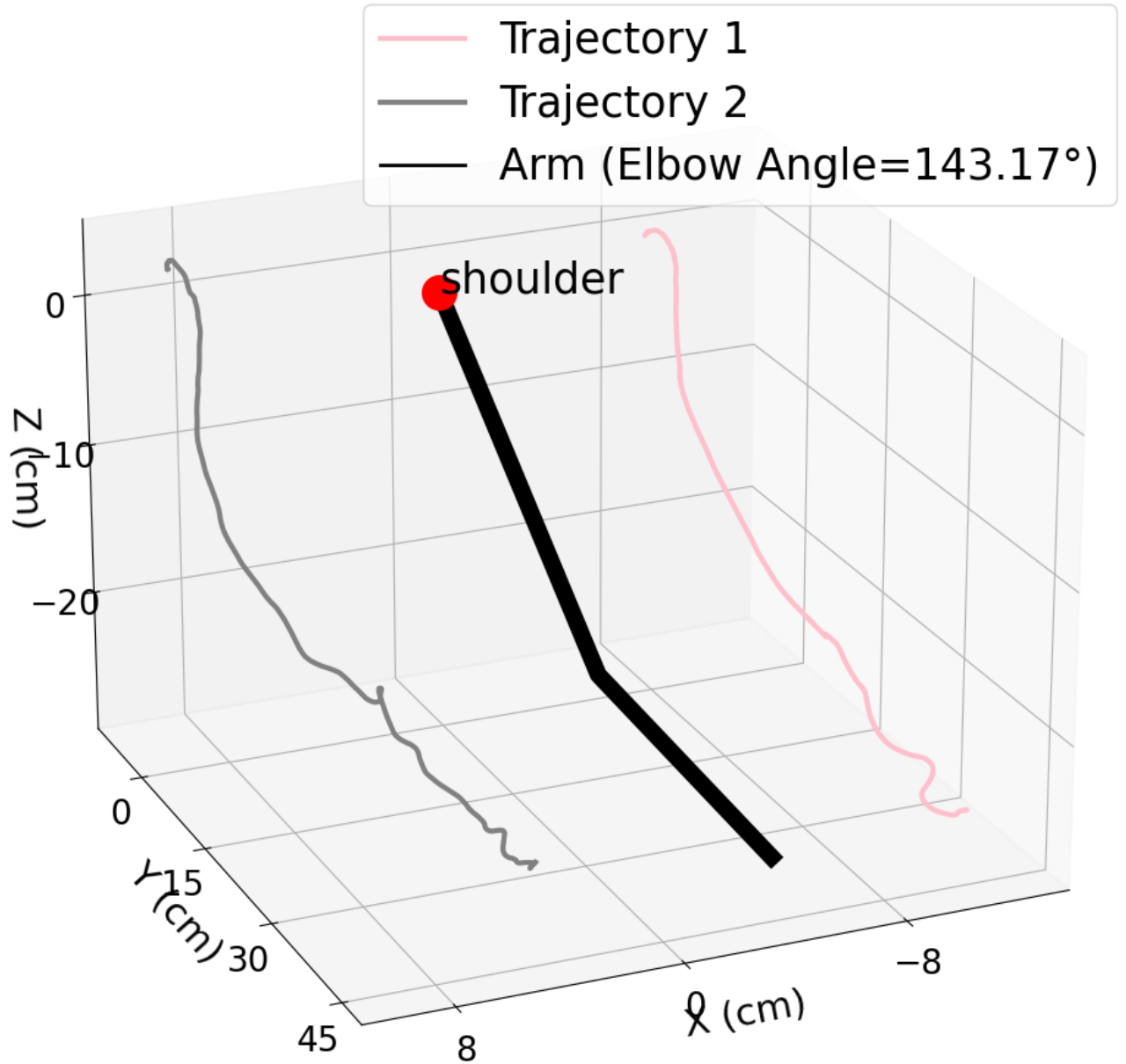}
      \caption{}
      \label{fig:data_process:pose_traj:traj_7}
    \end{subfigure}%
    \hfill
    \begin{subfigure}[b]{0.24\textwidth}
      \centering
      \includegraphics[width=\textwidth]{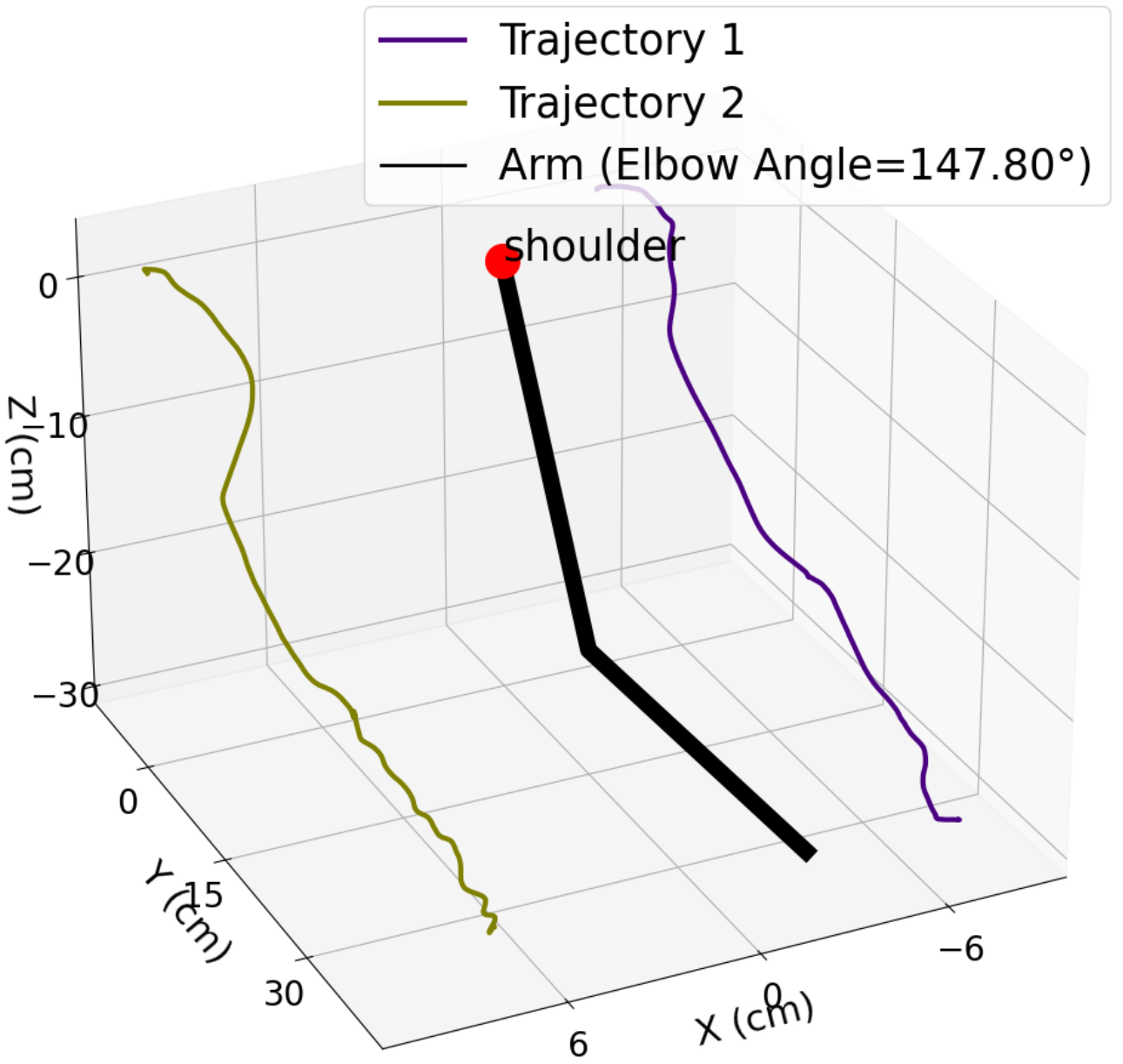}
      \caption{}
      \label{fig:data_process:pose_traj:traj_8}
    \end{subfigure}
  \end{minipage}
  \caption{Expert demonstrations on different postures used for training the dressing policy. The red dot represents the shoulder joint (a) Elbow angle \( \psi \) is \( 120.94^\circ \). (b) Elbow angle \( \psi \) is \( 128.80^\circ \). (c) Elbow angle \( \psi \) is \( 129.60^\circ \). (d) Elbow angle \( \psi \) is \( 131.87^\circ \). (e) Elbow angle \( \psi \) is \( 136.36^\circ \). (f) Elbow angle \( \psi \) is \( 138.63^\circ \). (g) Elbow angle \( \psi \) is \( 143.17^\circ \). (h) Elbow angle \( \psi \) is \( 147.80^\circ \).}
  \label{fig:data_process:pose_traj}
\end{figure*}

The collected data is converted into the spherical coordinate system. Fig. \ref{fig:data_process:pose_traj} displays the dressing trajectories of the dual robotic arms and the corresponding arm postures for eight instances of dressing clothing. For clarity, the coordinate origin is translated to the shoulder joint. 

\begin{figure}[htbp]
  \centering  
  \includegraphics[width=0.45\textwidth]{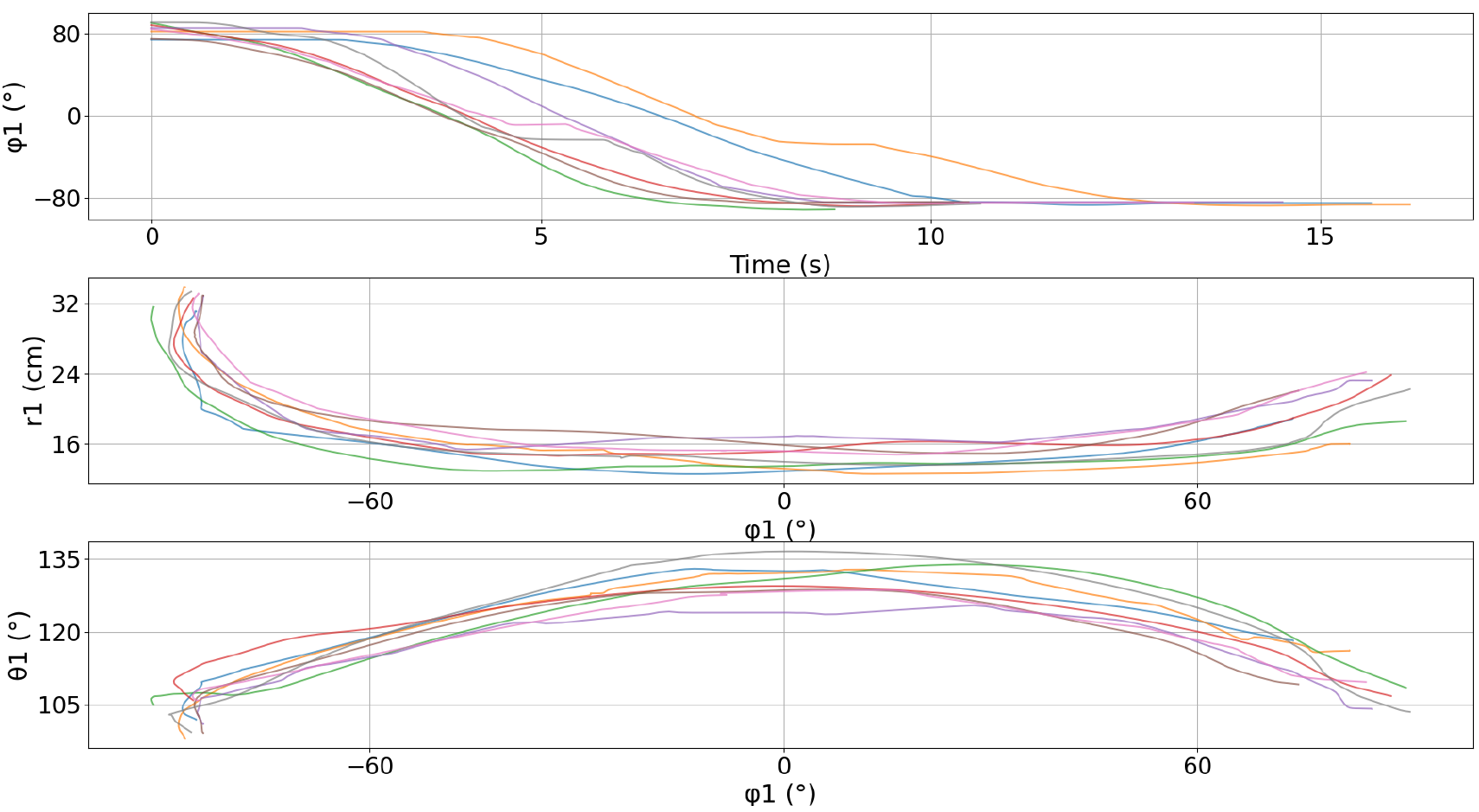}
  \caption{The eight trajectories of the first robotic arm in the spherical coordinate system during the dressing of tight clothing. (Top) \(\varphi_1\) over time. (Middle) \(r_1\) versus \(\varphi_1\). (Bottom) \(\theta_1\) versus \(\varphi_1\).}
  \label{fig:data_process:spher_time}
\end{figure}

In the data preprocessing stage, to reduce the impact of noise on the analysis results, this study employed the Locally Weighted Scatterplot Smoothing (LOWESS) method \cite{Cleveland01121979Robust} to smooth the raw data. Fig. \ref{fig:data_process:spher_time} displays the eight trajectories of the first robotic arm in the spherical coordinate system during the dressing of tight clothing. It can be observed that all the azimuthal angles start near \( \pi/2 \) and decrease monotonically over time until they reach values near \( -\pi/2 \).

\begin{figure}[htbp]
  \centering %
  \includegraphics[width=0.45\textwidth]{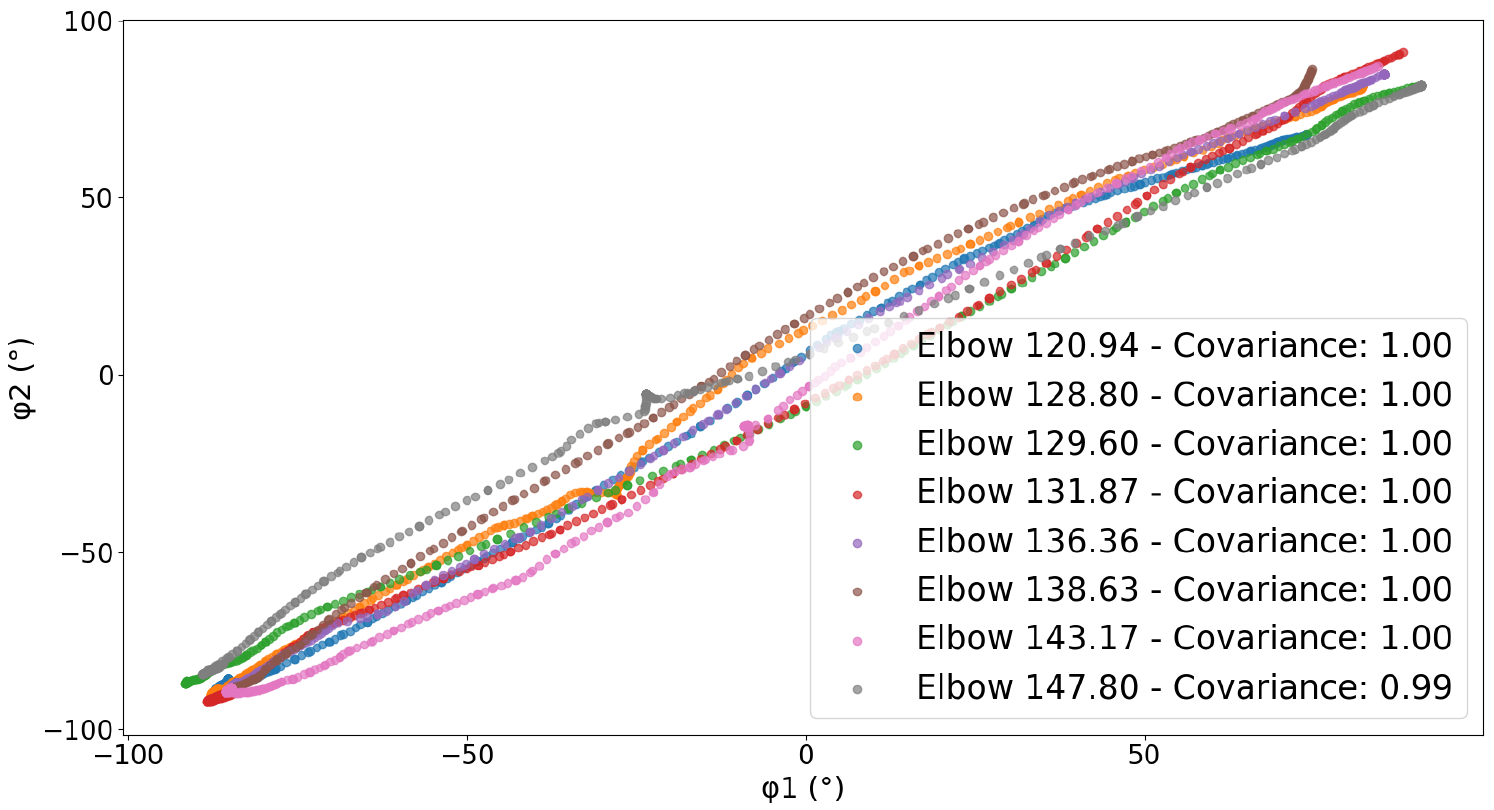} %
  \caption{The relationship (Covariance) between \(\varphi_1\) and \(\varphi_2\) variations of the two robotic arms across eight trials.} %
  \label{fig:data_process:psi_1_psi_2} %
\end{figure}

Fig. \ref{fig:data_process:psi_1_psi_2} shows the relationship between \( \psi_1 \) and \( \psi_2 \) at eight different elbow joint angles, along with the covariance of the eight curves. It can be seen that \( \psi_1 \) and \( \psi_2 \) exhibit a strong linear correlation, which quantitatively verifies the inference of the task-relevant features of the two curves proposed in Section \ref{sec:method}.

\begin{figure}[htbp]
    \centering

    \begin{subfigure}{0.23\textwidth}
        \centering
        \includegraphics[width=\textwidth]{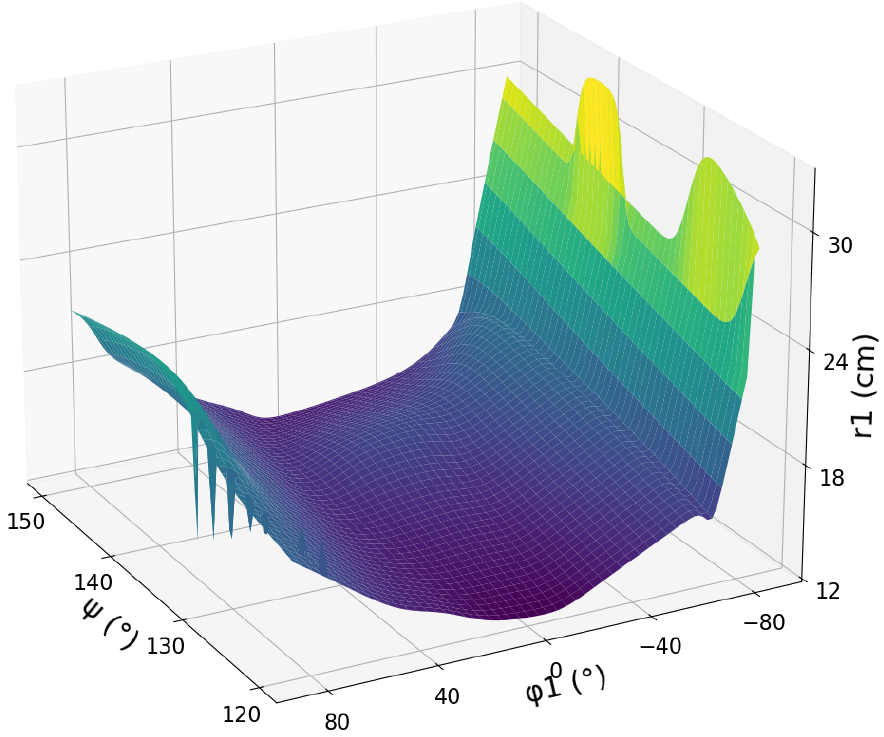} 
        \caption{}
    \end{subfigure} \hfill
    \begin{subfigure}{0.23\textwidth}
        \centering
        \includegraphics[width=\textwidth]{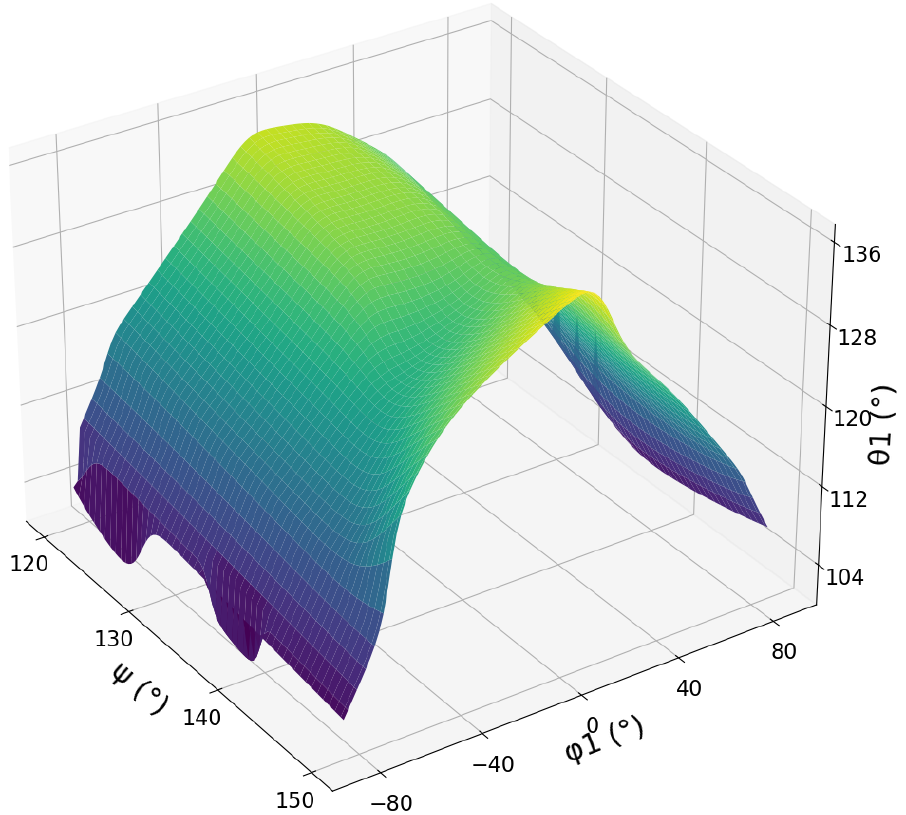} 
        \caption{}
    \end{subfigure}


    
    \vspace{1em}
    
    \begin{subfigure}{0.23\textwidth}
        \centering
        \includegraphics[width=\textwidth]{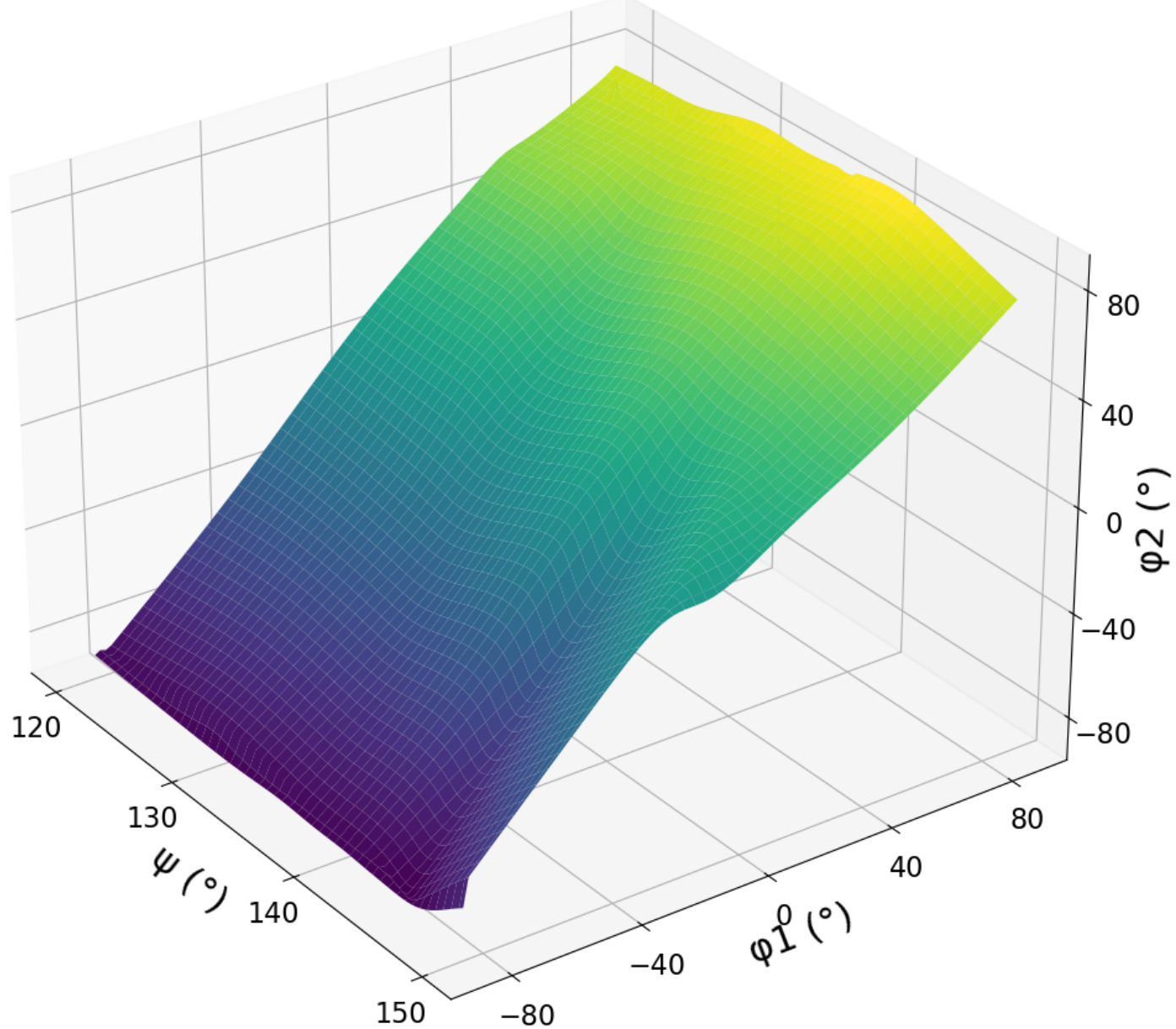}
        \caption{}
    \end{subfigure}

    \caption{The 3D surface plots for the first and the third GMMs. (a) The plots of the GMM for encoding \( r_1 \) and \( \theta_1 \), with \( \varphi_1 \) and \( \psi \) as inputs. (b) The plots of the GMM for encoding \( r_2 \) and \( \theta_2 \), with \( \varphi_2 \) and \( \psi \) as inputs. (c) The plot of the GMM for encoding \( \varphi_2 \), with \( \varphi_1 \) and \( \psi \) as inputs.}
    \label{fig:data_train:r1_theta_phi2}
    
\end{figure}

Three GMMs were trained. The first GMM takes as input (\( \varphi_1 \), \( \psi \)) and outputs \((r_1, \theta_1)\). The second GMM takes as input \((\varphi_2, \psi)\) and outputs \((r_2, \theta_2)\). The third GMM takes as input (\( \varphi_1 \), \( \psi \)) and outputs \( \varphi_2 \). The results are presented in the 3-D surface plot in Fig. \ref{fig:data_train:r1_theta_phi2}.

\subsection{Task Testing}
\label{subsec:experiment:task_test}
The trained GMM is applied to real-world experiments, four experiments were conducted:
\begin{enumerate}
    \item Bimanual dressing with loose clothes
    \item Bimanual dressing with tight clothes
    \item Single robotic arm dressing with loose clothes
    \item Single robotic arm dressing with tight clothes
\end{enumerate}

For each bimanual dressing experiment, different human arm postures were tested. First, the joint positions using the robotic arm's proprioceptive sensors were measured, then the corresponding \(\psi\) and \( \varphi_1 \) were calculated. Next, the trajectories of the two robotic arms based on the trained GMM were generated. For the single robotic arm experiments, only the first generated trajectory was used. Although the collected data only comes from tight garments, the trained GMMs can be applied to experiments with loose garments. For tight garments, 12 experiments were conducted for both single-arm and bimanual strategies, while for loose garments, 8 experiments were conducted at similar elbow joint angles. The test results are shown in Fig. \ref{fig:task_test:real_world_test}, and the full dressing sequences are recorded in the video (see Supplementary Materials). The experimental results are shown in Table \ref{tab:task_test:result_compare_rate}.

As shown in Fig. \ref{fig:task_test:real_world_test}, the bimanual dressing strategy has performed well on both tight and loose garments. However, the single-arm dressing strategy, while able to bring one side of the garment close to the shoulder joint, often fails to fully reach the shoulder joint on the other side due to the influence of stiffness decay (deformability). 

To evaluate the dressing, we introduce the concept of a \textquotedblleft dressing effectiveness indicator\textquotedblright. First, we define the upper arm vector \( \mathbf{v}_{\text{upper}} \) (the green line), then determine the average position of the armscye, \( \mathbf{P}_{\text{armscye}} \) (the blue line). Then, we calculate the projection of the vector from the elbow joint \( \mathbf{P}_{\text{elbow}} \) to \( \mathbf{P}_{\text{armscye}} \) onto \( \mathbf{v}_{\text{upper}} \), and compute the ratio of this projection to the length of the upper arm. This ratio serves as the dressing effectiveness indicator. The larger the ratio, the garment covers larger portion of the arm thus the dressing outcome is better. 

As shown in Table \ref{tab:task_test:result_compare_rate}, bimanual dressing results in larger dressing effectiveness, which shows a clear advantage over single-arm dressing when dressing with tight garments, ultimately achieving better dressing outcomes. 

\begin{figure*}[htbp]
  \centering
  
  \begin{minipage}[b]{0.16\textwidth}
    \centering
    \begin{subfigure}[b]{\textwidth}
      \includegraphics[width=\textwidth]{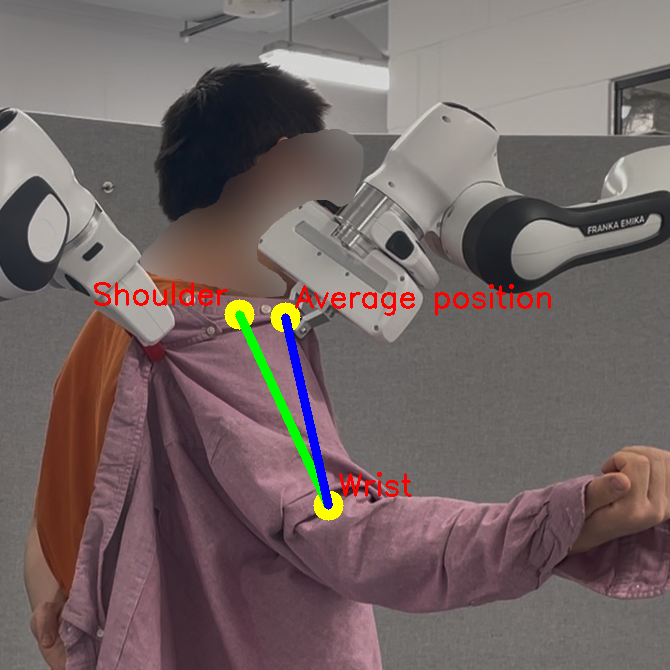}
      \caption{Start 1}
      \label{fig:task_test:real_world_test:1}
    \end{subfigure}
  \end{minipage}
  \hfill
  \begin{minipage}[b]{0.16\textwidth}
    \centering
    \begin{subfigure}[b]{\textwidth}
      \includegraphics[width=\textwidth]{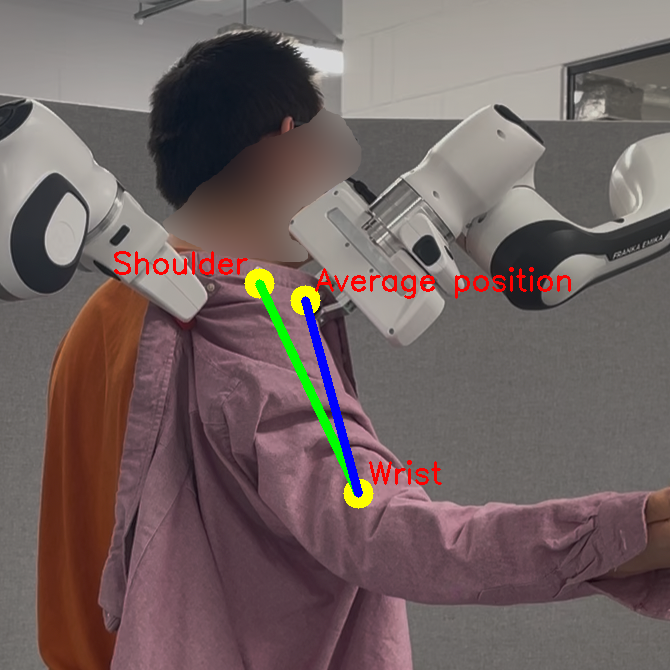}
      \caption{Start 2}
      \label{fig:task_test:real_world_test:2}
    \end{subfigure}
  \end{minipage}
  \hfill
  \begin{minipage}[b]{0.16\textwidth}
    \centering
    \begin{subfigure}[b]{\textwidth}
      \includegraphics[width=\textwidth]{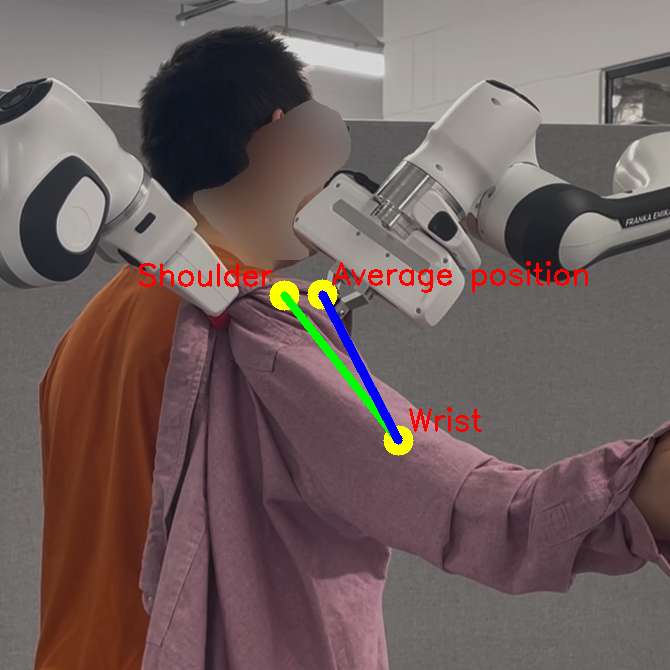}
      \caption{Start 3}
      \label{fig:task_test:real_world_test:3}
    \end{subfigure}
  \end{minipage}
  \hfill
  \begin{minipage}[b]{0.16\textwidth}
    \centering
    \begin{subfigure}[b]{\textwidth}
      \includegraphics[width=\textwidth]{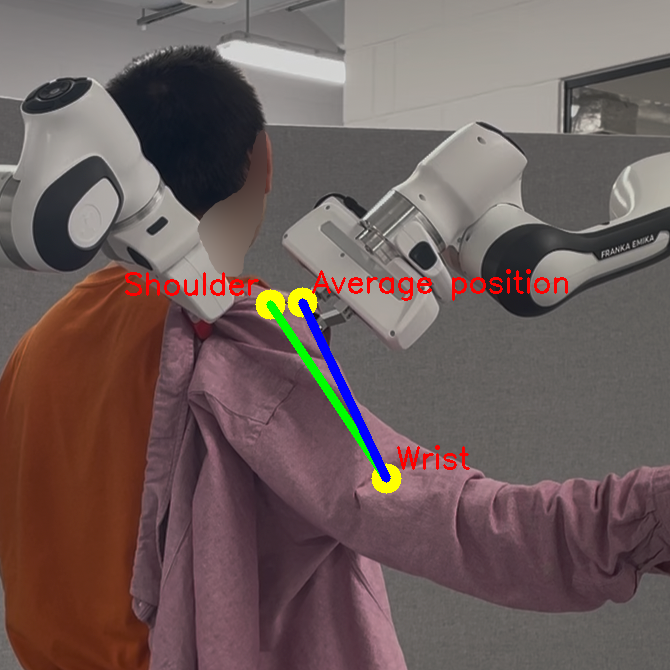}
      \caption{Start 4}
      \label{fig:task_test:real_world_test:4}
    \end{subfigure}
  \end{minipage}
  \hfill
  \begin{minipage}[b]{0.16\textwidth}
    \centering
    \begin{subfigure}[b]{\textwidth}
      \includegraphics[width=\textwidth]{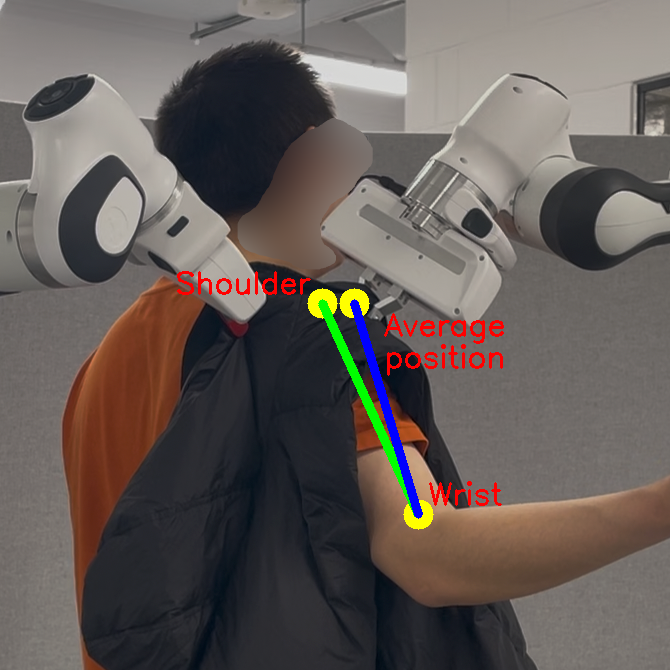}
      \caption{Start 5}
      \label{fig:task_test:real_world_test:5}
    \end{subfigure}
  \end{minipage}
  \hfill
  \begin{minipage}[b]{0.16\textwidth}
    \centering
    \begin{subfigure}[b]{\textwidth}
      \includegraphics[width=\textwidth]{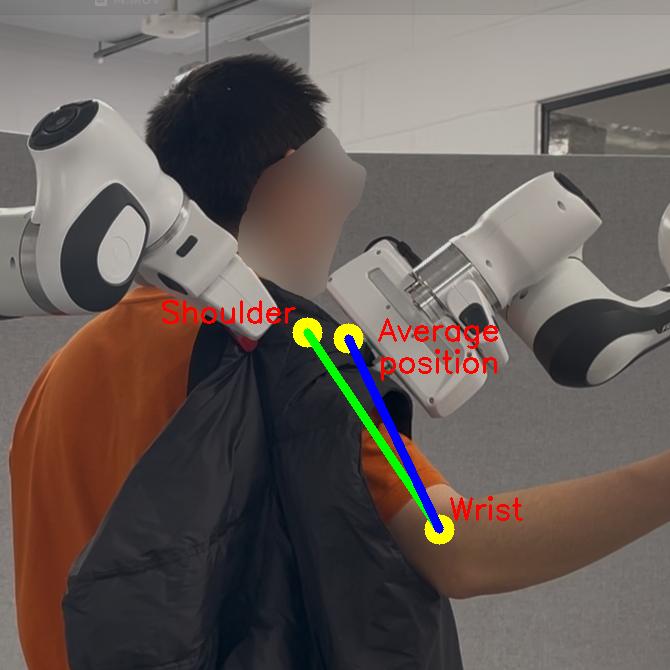}
      \caption{Start 6}
      \label{fig:task_test:real_world_test:6}
    \end{subfigure}
  \end{minipage}

  \vskip\baselineskip 
  
  \begin{minipage}[b]{0.16\textwidth}
    \centering
    \begin{subfigure}[b]{\textwidth}
      \includegraphics[width=\textwidth]{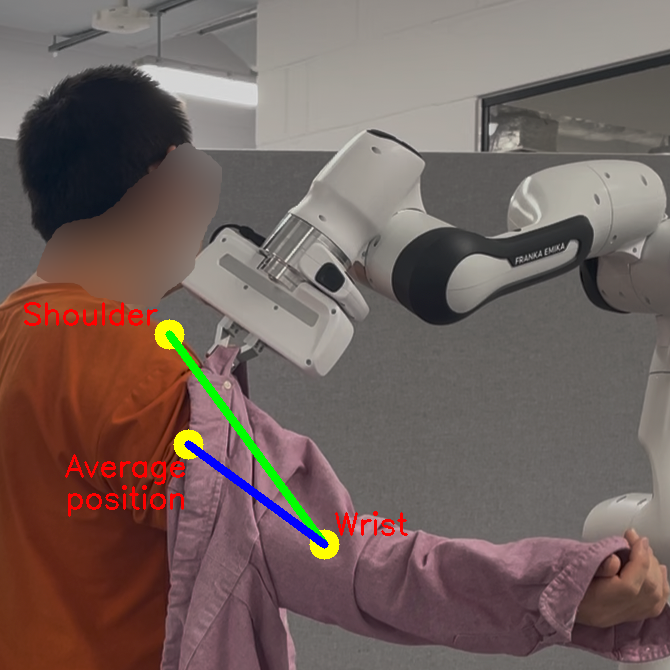}
      \caption{End 7}
      \label{fig:task_test:real_world_test:7}
    \end{subfigure}
  \end{minipage}
  \hfill
  \begin{minipage}[b]{0.16\textwidth}
    \centering
    \begin{subfigure}[b]{\textwidth}
      \includegraphics[width=\textwidth]{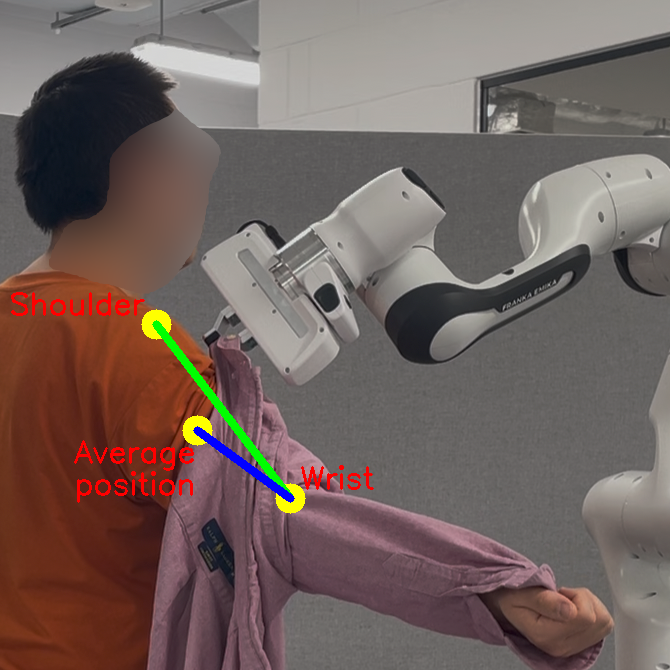}
      \caption{End 8}
      \label{fig:task_test:real_world_test:8}
    \end{subfigure}
  \end{minipage}
  \hfill
  \begin{minipage}[b]{0.16\textwidth}
    \centering
    \begin{subfigure}[b]{\textwidth}
      \includegraphics[width=\textwidth]{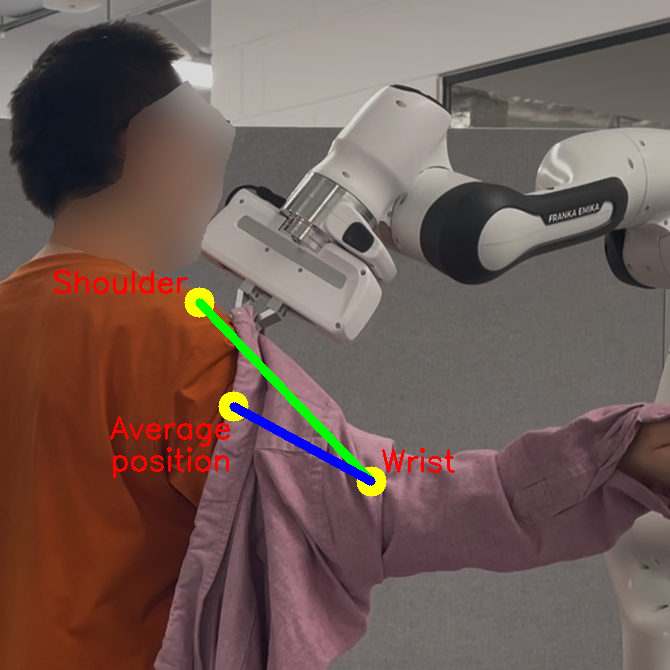}
      \caption{End 9}
      \label{fig:task_test:real_world_test:9}
    \end{subfigure}
  \end{minipage}
  \hfill
  \begin{minipage}[b]{0.16\textwidth}
    \centering
    \begin{subfigure}[b]{\textwidth}
      \includegraphics[width=\textwidth]{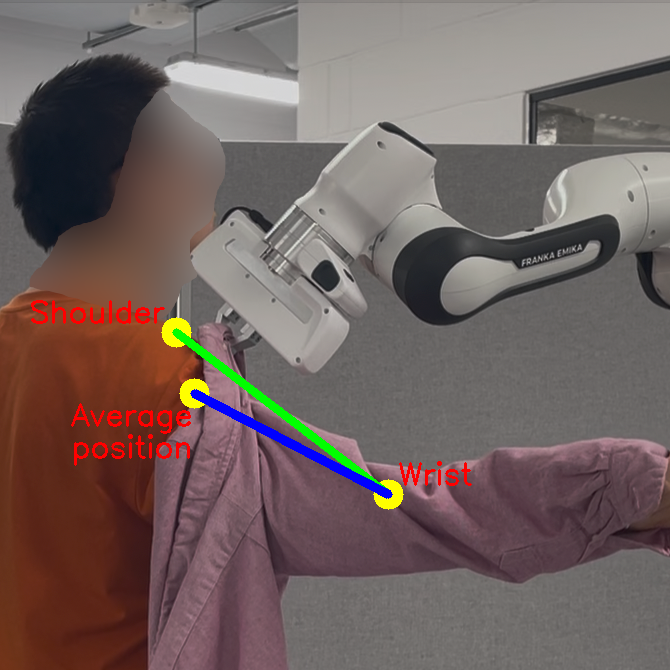}
      \caption{End 10}
      \label{fig:task_test:real_world_test:10}
    \end{subfigure}
  \end{minipage}
  \hfill
  \begin{minipage}[b]{0.16\textwidth}
    \centering
    \begin{subfigure}[b]{\textwidth}
      \includegraphics[width=\textwidth]{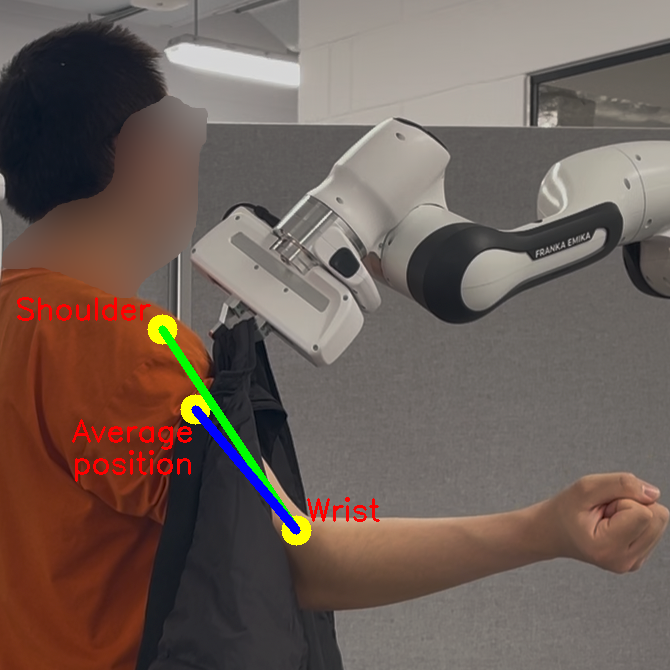}
      \caption{End 11}
      \label{fig:task_test:real_world_test:11}
    \end{subfigure}
  \end{minipage}
  \hfill
  \begin{minipage}[b]{0.16\textwidth}
    \centering
    \begin{subfigure}[b]{\textwidth}
      \includegraphics[width=\textwidth]{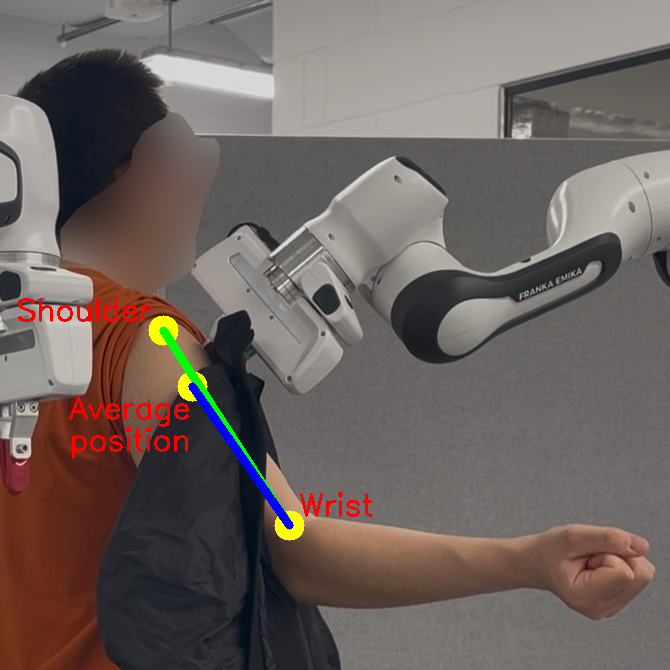}
      \caption{End 12}
      \label{fig:task_test:real_world_test:12}
    \end{subfigure}
  \end{minipage}

  \caption{Display of experimental results. The blue line represents the average position of the armscye on the upper arm, while the green line represents the entire upper arm. The ratio of their projections represents the dressing effect. (a-d) Bimanual dressing with tight garments. (e-f) Bimanual dressing with loose garments. (g-j) Single-arm dressing with tight garments. (k-l) Single-arm dressing with loose garments.}
  \label{fig:task_test:real_world_test}
\end{figure*}

\begin{table}[htbt]
\caption{Comparison of Dual and Single Measurements for Tight and Loose Garments. \(\psi\) represents the elbow joint angle, and Res. represents the percentage value of the dressing effectiveness indicator, where 0\% indicates that the garment has fallen off.} 
\begin{tabular}{llllllll}
\hline
\multicolumn{1}{c}{\multirow{3}{*}{No.}} & \multicolumn{4}{c}{Tight Garment}                                                                                                                     & \multicolumn{3}{c}{Loose Garment}                                                                                                                                                                                                                                    \\ \cline{2-8} 
\multicolumn{1}{c}{}                     & \multicolumn{2}{c}{Dual}                                              & \multicolumn{2}{c}{Single}                                                & \multicolumn{1}{c}{\multirow{2}{*}{\(\psi\)}} & \multicolumn{1}{c}{\multirow{2}{*}{\begin{tabular}[c]{@{}c@{}}Dual\\ Res.\end{tabular}}} & \multicolumn{1}{c}{\multirow{2}{*}{\begin{tabular}[c]{@{}c@{}}Single\\ Res.\end{tabular}}} \\ \cline{2-5}
\multicolumn{1}{c}{}                     & \multicolumn{1}{c}{\(\psi(^\circ)\)} & \multicolumn{1}{c}{Res.(\%)} & \multicolumn{1}{c}{\(\psi\)} & \multicolumn{1}{c}{Res.} & \multicolumn{1}{c}{}                                   & \multicolumn{1}{c}{}                                                                                  & \multicolumn{1}{c}{}                                                                                \\ \hline
1                                        & \(125.7\)                             & \textbf{\(89\)}                   & \(130.9\)                             & \(62\)                            & \(136.2\)                                              & \(94\)                                                                                                & \(71\)                                                                                              \\ \hline
2                                        & \(137.4\)                             & \(100\)                           & \(132.8\)                             & \(61\)                            & \(129.5\)                                              & \(89\)                                                                                                & \(58\)                                                                                              \\ \hline
3                                        & \(132.5\)                             & \(90\)                            & \(141.8\)                             & \(66\)                            & \(139.7\)                                              & \(92\)                                                                                                & \(64\)                                                                                              \\ \hline
4                                        & \(139.6\)                             & \(84\)                            & \(141.9\)                             & \(59\)                            & \(130.4\)                                              & \(88\)                                                                                                & \(52\)                                                                                              \\ \hline
5                                        & \(138.9\)                             & \(95\)                            & \(140.3\)                             & \(50\)                            & \(129.5\)                                              & \(92\)                                                                                                & \(58\)                                                                                              \\ \hline
6                                        & \(120.3\)                             & \(97\)                            & \(139.8\)                             & \(0\)                             & \(142.6\)                                              & \(100\)                                                                                               & \(72\)                                                                                              \\ \hline
7                                        & \(147.3\)                             & \(96\)                            & \(143.1\)                             & \(71\)                            & \(141.5\)                                              & \(100\)                                                                                               & \(64\)                                                                                              \\ \hline
8                                        & \(129.8\)                             & \(87\)                            & \(125.6\)                             & \(53\)                            & \(132.2\)                                              & \(95\)                                                                                                & \(72\)                                                                                              \\ \hline
9                                        & \(144.9\)                             & \(96\)                            & \(124.9\)                             & \(70\)                            &                                                        &                                                                                                       &                                                                                                     \\ \cline{1-5}
10                                       & \(132.8\)                             & \(92\)                            & \(127.0\)                             & \(68\)                            &                                                        &                                                                                                       &                                                                                                     \\ \cline{1-5}
11                                       & \(134.1\)                             & \(93\)                            & \(150.6\)                             & \(67\)                            &                                                        &                                                                                                       &                                                                                                     \\ \cline{1-5}
12                                       & \( 142.2 \)                           & \(96\)                            & \(137.3\)                             & \(80\)                            &                                                        &                                                                                                       &                                                                                                     \\ \cline{1-5}
\end{tabular}
\label{tab:task_test:result_compare_rate}
\end{table}

\section{CONCLUSIONS}
\label{sec:conclusion}
Traditional robot-assisted dressing focuses on how to dress with loose clothing, where a single robotic arm is often sufficient to complete the task. However, when dressing with tight clothing using a single robotic arm, failures frequently occur due to the narrower armscye and the property of diminishing rigidity. To address the issue of dressing with tight clothes, this paper proposes a bimanual assisted dressing strategy. To implement this strategy, a spherical coordinate system is established to facilitate encoding and identify task-relevant features for bimanual assistance, and then use GMM/GMR imitation learning to replicate the human bimanual dressing strategy. Finally, the effectiveness of the proposed method is validated through various experiments.

Nevertheless, several important directions remain for future work. First, the generalization capability of the proposed method across varying human arm lengths and joint configurations should be further evaluated. Second, a comparative analysis between GMM and alternative models such as Dynamic Movement Primitives (DMP), Gaussian Processes, and neural network-based policies is necessary to assess the relative strengths and weaknesses. Third, robustness under slight arm movements and garment deformations should be experimentally studied to better reflect real-world conditions. These extensions will help validate and enhance the practical applicability of the proposed framework for robot-assisted dressing with tight garments. Moreover, the current method assumes that the human arm remains static during dressing, which may not hold for elderly individuals. Therefore, future research is needed to explore dynamic dressing strategies that can adapt to human arm motion and involuntary movements

Despite the limitation above, the research presented in this paper provides a new perspective for robotic collaborative dressing, particularly for tight clothing. In the future, further optimization of the coordinate system will be investigated and imitation learning algorithm and continue experimenting with tight garments such as socks, gloves, and thermal pants.


\section*{ACKNOWLEDGMENT}
We sincerely appreciate John Bateman for his valuable suggestions on the spherical coordinate system.

\bibliographystyle{ieeetr}
\bibliography{references}

\end{document}